\documentclass[sigconf]{acmart}
\usepackage{subcaption}
\usepackage{enumitem}

\AtBeginDocument{%
  }

\copyrightyear{2026}
\acmYear{2026}
\setcopyright{cc}
\setcctype{by}
\acmConference[MM '26] {Proceedings of the 34th ACM International Conference on Multimedia}{November 10--14, 2026}{Rio de Janeiro, Brazil.}
\acmBooktitle{Proceedings of the 34th ACM International Conference on Multimedia (MM '26), November 10--14, 2026, Rio de Janeiro, Brazil}
\acmISBN{979-8-4007-2213-4/2026/11}

\setcopyright{acmlicensed}
\copyrightyear{2026}
\acmYear{2026}
\setcopyright{cc}
\setcctype{by}
\acmDOI{10.1145/3767308.3836396}
\acmConference[MM '26] {Proceedings of the 34th ACM International Conference on Multimedia}{November 10--14, 2026}{Rio de Janeiro, Brazil.}
\acmBooktitle{Proceedings of the 34th ACM International Conference on Multimedia (MM '26), November 10--14, 2026, Rio de Janeiro, Brazil}
\acmISBN{979-8-4007-2213-4/2026/11}
\acmISBN{979-8-4007-2213-4/2026/11}

\begin{document}

\title{Geometry-guided Emotion Modulation for Controllable and Photorealistic Emotional Talking Face Generation}

\author{Chenggong Hu}
\authornote{Both authors contributed equally to this research.}
\email{huchenggong@zju.edu.cn}
\orcid{0009-0004-4792-2042}
\affiliation{%
  \institution{School of Software Technology, Zhejiang University}
  \city{Ningbo}
  \country{China}
}

\author{Shaoyin Ma}
\authornotemark[1]
\email{mashaoyin@zju.edu.cn}
\orcid{0009-0000-3045-6111}
\affiliation{%
  \institution{School of Software Technology, Zhejiang University}
  \city{Ningbo}
  \country{China}
}

\author{Yi Wang}
\email{y\_w@zju.edu.cn}
\orcid{0009-0006-9396-9874}
\affiliation{%
  \institution{College of Computer Science and Technology, Zhejiang University}
  \city{Hangzhou}
  \country{China}
}

\author{Li Sun}
\email{lsun@zju.edu.cn}
\orcid{0000-0001-8415-6579}
\affiliation{%
  \institution{Ningbo Global Innovation Center, Zhejiang University}
  \city{Ningbo}
  \country{China}
}

\author{Mingli Song}
\email{brooksong@zju.edu.cn}
\orcid{0000-0003-2621-6048}
\affiliation{%
  \institution{College of Computer Science and Technology, Zhejiang University}
  \city{Hangzhou}
  \country{China}
}

\author{Jie Song}
\authornote{Corresponding author.}
\email{sjie@zju.edu.cn}
\orcid{0000-0003-3671-6521}
\affiliation{%
  \institution{School of Software Technology, Zhejiang University}
  \city{Ningbo}
  \country{China}
}

\renewcommand{\shortauthors}{Chenggong Hu et al.}

\begin{abstract}
  Audio-driven emotional talking face generation aims to synthesize realistic videos with expressive facial dynamics. However, existing methods struggle to balance controllability and visual fidelity. Although implicit representations capture rich semantics, they lack structural guidance, often resulting in averaged emotional expressions. In contrast, explicit geometric methods offer better control over facial expressions but tend to sacrifice high-frequency texture details. To address it, we propose GemTalk, a diffusion-based framework that combines the semantic richness of implicit representations with the structural precision of explicit geometric priors. We introduce a Vision-guided Audio Emotion Projection (V-AEP) module to extract implicit emotional lip and expression features. At the same time, a Diffusion-based Geometric Priors Generator (D-GPG) generates identity-aware blendshape coefficients as explicit structural priors. Crucially, our Geometry-guided Emotion Modulation (GEM) module leverages these geometric priors to recalibrate the magnitude of implicit features, enabling precise, continuous control over emotional expressions, especially emotion intensity, without sacrificing visual quality. Extensive experiments show GemTalk achieves superior performance in photo-realism, and facial emotional dynamics.
\end{abstract}

\ccsdesc[500]{Computing methodologies~Computer vision problems}
\ccsdesc[300]{Computing methodologies~Reconstruction}
\ccsdesc[100]{Computing methodologies~Image representations}

\keywords{diffusion model, talking face generation, representation learning}


\maketitle

\begin{figure}[h]
  \centering
  \includegraphics[width=1\linewidth]{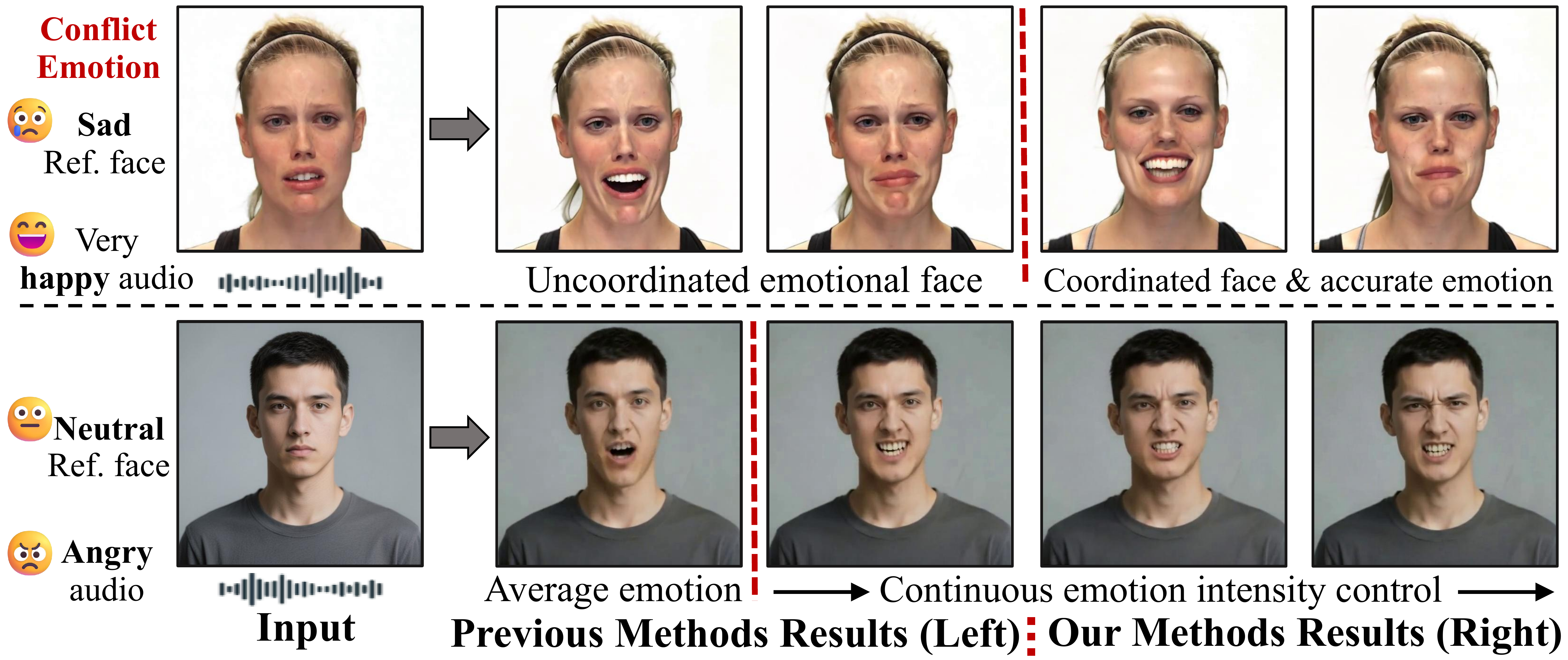}
  \caption{Top: Explicit driving methods (left) yield uncoordinated faces, while our fusion method (right) produces harmonious appearances. Bottom: Implicit generation methods (left) lack continuous control over emotion details, while ours (right) enables continuous geometric editing.}
  \label{fig:intro_pic}
\end{figure}

\section{Introduction}

Talking face generation aims to create realistic and expressive videos from a reference face and audio, with applications in virtual avatars, gaming, and filmmaking. Recent works \cite{wav2lip,diff2lip,musetalk,speechdriven,lipexpert,text2lip}, particularly diffusion-based models \cite{musetalk,speechdriven}, have achieved accurate lip synchronization (sync), but upper facial expression is often overshadowed, as the dominant audio-articulation correlation suppresses subtle emotional cues from audio, resulting neutral or rigid face. To address this, previous works \cite{dicetalk,sadtalker,aniportrait} decouple facial expression from lip sync and utilize different representations from audio, video, or text to do generation. These approaches generally fall into two categories: those \cite{hallo,sonic} utilizing \textbf{implicit representation} to directly \textbf{guide the synthesis} of facial expressions, and those employing \cite{EAT,styletalk} \textbf{explicit geometric representation} (e.g., 3D coefficients \cite{3dmm,3dlatent}, blendshape coefficients \cite{blendshape}, or learned motion bases \cite{edtalk} to \textbf{drive or render} the facial animation.

For implicit representation, it can guide powerful generative models, such as diffusion models, to synthesize high-quality videos, but suffers from inherent controllability limitations. Hallo \cite{hallo} captures only audio prosody rather than semantic emotion, leading to ambiguity where states like loudness are indistinguishable from anger. Although MEMO \cite{memo} and EmoTalker \cite{emotalker} extract emotional features from audio or text, and DICE-Talk \cite{dicetalk} and EmotiveTalk \cite{emotivetalk} further enhance expression by constructing an emotional audio-visual features, they all lack explicit structural guidance. Consequently, these representations fail to perceive specific geometric deformation magnitude. This deficiency manifests as uncontrollable intensity and uncoordinated facial dynamics in emotion modeling as shown in Figure \ref{fig:intro_pic}. Even with discrete intensity supervision, the signal remains a discrete numeral devoid of fine geometric nuance, resulting in averaged emotional expressions. Furthermore, due to the inability of these coarse-grained representations to precisely drive facial muscles, they remain susceptible to the emotion interference from the face image. Conversely, for explicit geometric representation, it prioritizes controllability but often compromise visual fidelity. SadTalker\cite{sadtalker}, StyleTalk \cite{styletalk}, and EAT \cite{EAT} rely on warping reference faces, frequently causing severe pixel distortion, leading strange facial distortions and identity loss. EDTalk \cite{edtalk} and EmoTalk \cite{emotalk} are limited by linear motion bases or blendshap coefficients rendering, failing to synthesize photorealistic high-frequency textures. 

Therefore, achieving precise emotional control without compromising high visual fidelity necessitates synergizing the semantic richness of implicit representations with the structural precision of explicit signals. Recently, EchoMimic \cite{echomimic} combines these modalities by treating landmarks as spatial conditions alongside implicit representations. However, it fundamentally imposes geometry as a rigid spatial constraint. This mechanical combination\cite{continuationtrustworthycoarsescale} ignores the intrinsic correlation between emotional semantics and physical deformations, failing to achieve the deep synergy required for vivid and coordinated facial dynamics.

To synergize the generative fidelity of implicit representations with the structural precision of explicit geometric signals for controllable and photorealistic emotional talking face generation, we propose GemTalk, a diffusion-based framework employing a progressive refinement strategy. We first establish a emotion-agnostic lip sync backbone trained on large-scale non-emotional data, providing a stable foundation for subsequent emotional modeling. Then, to extract robust implicit emotional representations, we propose a Vision-guided Audio Emotion Projection (V-AEP) framework. The core objective of V-AEP is to bridge the inherent audio-visual modality gap, leveraging vision signals to guide to extract emotion-related lip and expression representations from audio that are specifically optimized for visual generation tasks. Simultaneously, a Lip-Exp (exp denotes expression in the paper) Coordination Module (LECM) is introduced to model the synergistic interaction between lip and expression, ensuring facial integrity. To obtain explicit geometric signals, we propose Diffusion-based Geometric Priors Generator (D-GPG) to synthesize identity-aware blendshape coefficients from audio. Finally we employ a novel Geometry-guided Emotion Modulation (GEM) module inspired by the geometric interpretation of the cross attention inner product to integrate the strengths of them. In diffusion models\cite{semi,mamba,seeingendstepzero,resilphase}, the conditions are injected relying on the cross attention to achieve conditional generation. Geometrically, the attention score is governed by the inner product $\|Q\| \cdot \|K\| \cdot \cos\theta$. In our task, the guidance is driven not only by the feature direction ($\cos\theta$, determining emotion category), but critically by the magnitude ($\|K||$, encoding the physical intensity manifested as specific facial component deformations). However, We observe that while \textbf{implicit representation} accurately capture the emotion direction, their magnitude lacks structural grounding, often exhibiting uninformative variance or intensity collapse, which prevents the model from controlling facial structural details, especially the emotional intensity. This phenomenon mirrors the motivation behind RMSNorm, which has played a pivotal role in stabilizing modern Large Language Models (LLMs) by decoupling feature direction from magnitude. However, while RMSNorm effectively mitigates variance, it relies on static, data-independent parameters for scaling, making it insufficient for modeling dynamic, fine-grained physical expressions. Advancing this normalization philosophy into the physical domain, our GEM utilizes explicit geometric representations as structural priors to dynamically recalibrate the implicit feature magnitude. By injecting geometric cues without altering the semantic direction, GEM ensures that the cross-attention mechanism receives guidance enriched with precise, instance-specific physical intensity. This synergy endows the implicit features with global geometric awareness, allowing the model to accurately perceive plausible distributions of facial muscle activations. Consequently, it achieves anatomically coordinated facial dynamics with precise muscle regulation, specifically enabling continuous control over emotional intensity. \textit{Theoretical analysis of uncontrollable magnitude and its impact on attention mechanism in diffusion model are provided in the Appendix.}
%

In summary, our main contributions are as follows: (1) We propose GemTalk, a pure audio-driven diffusion framework that bridges implicit generation and explicit driving to achieve high-quality, controllable emotional talking face generation, orchestrating rich facial emotions from simple audio without relying on complex auxiliary inputs like driving videos; (2) We design GEM based on the geometric nature of cross attention. It endows the representations with facial structural awareness by utilizing geometric signals to recalibrate feature magnitude, enabling continuous emotional intensity control without discrete labels; (3) we design V-AEP and LECM to extract decoupled but synergistic lip and expression emotion features from audio; (4) we propose D-GPG to synthesize accurate and identity-aware blendshape coefficients from audio, providing geometric priors for intensity modulation.

\section{Related Work}
\begin{figure*}[t]
  \centering
  \includegraphics[width=1 \textwidth, trim=0 0 0 0, clip]{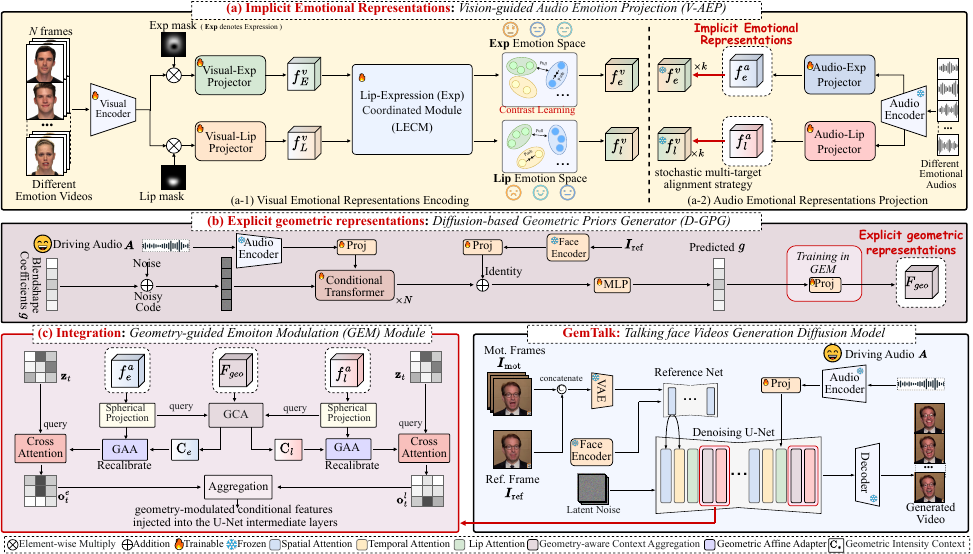}
  \caption{Given a reference face and driving audio, our GemTalk generates high-fidelity and controllable emotional videos. We first establish a emotion-agnostic lip sync backbone. Then we employ the V-AEP framework to extract robust implicit emotional representations, and D-GPG to synthesize explicit geometric representations. Finally, the GEM utilizes explicit geometric representations as structural priors to recalibrate the implicit feature magnitude, enabling continuous control over facial dynamics without losing visual quality. Details of the baseline framework are provided in the Section \ref{method}.}
  \label{fig:main_pic}
\end{figure*}

\textbf{Audio-Driven Talking Face Generation.} 
Early works \cite{lipexpert,videoretalking,musetalk,wav2lip,diff2lip} only focused on audio to lip mapping, keeping other facial attributes static. To drive facial expressions, AniPortrait \cite{aniportrait}, SadTalker \cite{sadtalker}, and MODA \cite{moda} use \textbf{explicit representations} like facial landmarks or 3D meshes to generate lip motion and facial dynamics. Similarly, AniTalker \cite{anitalker} explores an identity-decoupled implicit motion space from audio; however, its final synthesis still relies on an explicit renderer to warp a single reference image. Conversely, with the recent advancements in generative models, \textbf{implicit methods} like Hallo \cite{hallo}, Sonic \cite{sonic}, and MCDM \cite{mcdm} establish a direct conditional generation paradigm. They bypass explicit structural rendering, generating realistic talking face videos by directly injecting audio features into the implicit latent space of diffusion models. Although some recent diffusion-based frameworks, such as V-Express \cite{vexpress} and EchoMimic \cite{echomimic}, incorporate explicit spatial signals (e.g., facial landmarks) as auxiliary hard constraints to guide the generation, these explicit cues act merely as external driving conditions and are not deeply fused with the implicit representations. Consequently, while excelling in visual quality and lip sync, these audio-driven methods still fail to align expressions with the audio's affective tone, ultimately lacking the capacity for fine-grained, controllable emotional driving.

\noindent\textbf{Emotional talking face generation.} Early methods \cite{jiaudio,emotional} use discrete emotion labels for emotion modeling. More recently, some works \cite{EmoHead, emotalk, EAT} leverage 3D representations or blendshape coefficients to map audio to facial motion. However, these coarse-grained representations inherently constrain complex emotional diversity. Some works \cite{emotalker,mesl} achieve text-driven emotion control by aligning textual and visual representations within the CLIP space, but text often struggles with controlling complex facial expressions. To increase controllability, some models \cite{styletalk,eamm,PD-FCG,X-portrait,edtalk} utilize video references to transfer emotional dynamics while maintaining lip sync. However, requiring auxiliary driving videos inevitably complicates the input process and reduces overall convenience. To avoid such complex inputs, recent purely audio-driven approaches \cite{emotivetalk} attempt to extract visual emotional cues directly from the audio signal. By establishing a joint audio-visual space, they acquire expression-related representations, endowing the audio with visual semantic information to generate rich facial emotions. However, they all struggle with fine-grained intensity modulation and neglect facial coordination, resulting in strange faces. In our work, we use both explicit 3D blendshape coefficients and implicit emotional representations from audio with our GEM module to achieve plausible, intensity-controllable expressions without compromising video quality, identity and lip sync priors.

\section{Method}
\label{method}



\noindent\textbf{Baseline Framework.} The overall architecture of GemTalk is illustrated in Figure \ref{fig:main_pic}. We draw upon the design of the reference net from AnimateAnyone \cite{animate}, as well as the motion-frame strategy from Diffused-Heads \cite{diffused}, and CyberHost \cite{cyberhost}, to construct a baseline framework. Specifically, we use a parallel 2D UNet as a reference net to encode a reference image $\boldsymbol{I}\text{ref}$ to reference features and motion frames $\boldsymbol{I}\text{mot}$ to motion features. The denoising U-Net is based on the 2D UNet, integrating a pretrained temporal attention from AnimateDiff \cite{animatediff}, enabling it to predict facial video clips. Rather than employing computationally expensive 3D video backbones (e.g., 3D DiT/VAE), we deliberately adopt this 2D-plus-motion paradigm. Given the minimal temporal variance in talking heads, it drastically reduces training costs while fully preserving powerful pre-trained 2D spatial priors.

\noindent\textbf{Proposed designs.} As shown in Figure \ref{fig:main_pic}, we extract robust implicit emotional representations through V-AEP (Section \ref{sec:V-AEP}), then we employ D-GPG (Section \ref{sec:D-GPG}) to derive explicit geometric signals. Finally, the GEM (Section \ref{sec:GEM}) utilizes the explicit geometric signals to recalibrate the implicit representations, endowing the generation process with fine-grained structural awareness. We provide some information of the training and inference in section \ref{sec:TRAINING}.

\subsection{Vision-guided Audio Emotion Projection}
\label{sec:V-AEP}
Here, we details the extraction of emotion-aware and coordinated implicit lip and expression representations. We leverage emotional vision supervision to bridge the modality gap, rendering acoustic feature optimal for visual synthesis.

\noindent\textbf{Visual Emotional Representations Encoding.} This module aims to extract emotion-discriminative representation spaces from video frames while suppressing identity and background noise. Given a training batch of emotional videos $\boldsymbol{X}={\{x_i\}}^b_{i=1} \in \mathbb{R}^{B\times N\times c\times h\times w}$, each comprising $N$ randomly consecutive frames, we extract basic facial feature via a visual encoder \cite{resemotenet}. Then, we utilize lip and expression masks to isolate region-specific features and input them to the dedicated visual-lip and visual-exp projector to yield the initially embeddings $f_E^v$, $f_L^v\in\mathbb{R}^{B\times N\times d}$. To ensure facial integrity, we apply a Lip-Exp Coordination Module (LECM) to model the synergistic interaction between them. Specifically, taking expression as an example, we employ cross attention to retrieve relevant contextual cues from lip. $f_E^v$ serves as the query, and $f_L^v$ as the key-value pair to retrieve the expression-aware context, named $\boldsymbol{C}_l$, which aggregates facial expression dynamics that are temporally aligned with the current lip movements, establishing a semantic interaction. Subsequently, $\boldsymbol{C}_l$ is injected back into the original lip feature via a learnable gated residual connection, where a gating parameter $\alpha_{L}$ is initialized to a small value to restrict the information flow:
\begin{equation}
\label{eq:coordinated2}
\begin{aligned}
f_e^v = f_E^v + \text{Sigmoid}(\alpha_L) \cdot \boldsymbol{C}_l,
\end{aligned}
\end{equation}
where ${f}_e^v$ is the integrated expression feature. The design ensures it retains its semantics while being coordinated with the lip dynamics. 

To construct a visual space that purely encodes emotional semantics without identity leakage, we perform supervised contrastive learning with hard negative mining on the integrated features $f_e^v$ and $f_l^v$. Taking the expression branch as an example, for each anchor sample $i$, we define a positive set $P(i)$ sharing the same emotion label, and a hard negative set $\mathcal{N}_{hard}(i)$ containing the top-$K$ samples with different labels that exhibit the highest similarity to the anchor. The loss is computed over their union set $\mathcal{S}(i) = P(i) \cup \mathcal{N}_{hard}(i)$:
\begin{equation}
\mathcal{L}_\text{exp}^\text{v} = \sum_{i=1}^{N} \frac{-1}{|P(i)|} \sum_{p \in P(i)} \log \frac{\exp\left( f_e^v(i) \cdot f_e^v(p) / \tau \right)}{\sum_{k \in \mathcal{S}(i)} \exp\left( f_e^v(i) \cdot f_e^v(k) / \tau \right)},
\end{equation}
where $\tau$ is the temperature parameter. The lip branch follows the exact same procedure to compute $\mathcal{L}_\text{lip}^\text{v}$. The total visual objective is formulated as $\mathcal{L}_\text{visual} = \lambda_1 \mathcal{L}_\text{exp}^\text{v} + \lambda_2 \mathcal{L}_\text{lip}^\text{v}$. This process constructs two stable and coordinated emotion spaces, serving as pure visual supervisory signals for subsequent audio projection. \textit{Detailed data augmentations and hyperparameter settings are in the Appendix.}

\noindent\textbf{Audio Emotional Representations Projection.} Given a batch of input audio sequences $\boldsymbol{A}$ spanning diverse emotions, we first utilize a frozen Wav2Vec \cite{wav2vec} as an audio encoder to extract multi-scale audio feature. Then, we employ two parallel audio projectors to map the audio feature into the emotional audio-lip feature $f_l^a$ and audio-expression feature $f_e^a$, both $\in\mathbb{R}^{B\times N\times m\times d}$, where $m$ denotes the length of the learned audio context tokens, which aggregate the audio multi-scale information. To train these projectors, we first perform average pooling along the $m$ dimension of the audio features to strictly match the visual dimensions. However, optimizing directly with paired samples ignores the inherent visual diversity, where a single audio emotion corresponds to multiple facial dynamics. To address this, we propose a stochastic multi-target alignment strategy. Unlike rigid one-to-one mapping, for each audio sample, we randomly select $k$ visual anchors from the current batch that share the same emotion. The model is optimized by maximizing the average cosine similarity between the pooled audio features and the $k$ targets. This constraint prevents overfitting to identity-specific details and enforces the learning of robust, generalized implicit emotional semantics $f^a_l$ and $f^a_e$. Only using these features to guide the model can generate accurate facial emotion. \textit{The details of loss function is provided in the Appendix.}

\subsection{Diffusion-based Geometric Priors 
Generator}
\label{sec:D-GPG}
To provide physically interpretable geometric priors, we utilize blendshape coefficients, which structurally parameterize facial dynamics via 52 distinct muscle activations. Based on statistical analysis, we discard $7$ dimensions whose maximum activation values consistently remain below $0.05$, retaining $K=45$ effective dimensions denoted as ${g} \in \mathbb{R}^{B \times N \times K}$. To synthesize them, we propose D-GPG, a diffusion model conditioned on the driving audio $\boldsymbol{A}$. We observe that facial dynamics exhibit inherent identity dependency, where coefficient magnitudes vary across individuals for identical emotions. Drawing inspiration from the linear superposition principle of 3DMM, we incorporate identity features derived from the image $\boldsymbol{I}_\text{ref}$ into the generation process. This ensures the synthesized motions are physiologically consistent with the target subject. The denoising network $\mathrm{n}_\theta$ of D-GPG is optimized as follows: 
\begin{equation}
    \mathcal{L}_\text{gpg} = \mathbb{E}_{{g}_t, \mathrm{n}\sim\mathcal{N}(0,1), t}\big[\|\mathrm{n} - \mathrm{n}_\theta({g}_{t}, t,\boldsymbol{I}_\text{ref},\boldsymbol{A})\|_2^2\big],
\end{equation}
where $\mathrm{n}$ is the noise added to the latent ${g}_t$ at timestep $t$. The synthesized scalar coefficients will be mapped via a learnable projection layer into high-dimensional geometric tokens ${F}_\text{geo}\in\mathbb{R}^{B\times N\times K\times d}$ in the GEM. \textit{Details about blendshape and D-GPG are in the Appendix.}

\subsection{Geometry-guided Emotion Modulation}
\label{sec:GEM}
The GEM is designed to endow implicit representations with fine-grained geometric structural awareness, enabling precise emotion modulation across facial dynamics. By leveraging explicit blendshape coefficients as physical priors, GEM \textbf{recalibrates} the feature magnitude in two coordinated stages, allowing continuous and controllable emotional intensity without sacrificing visual quality.

\noindent\textbf{Geometry-aware Context Aggregatio (GCA).}  
To endow the implicit representation with physical interpretability, we must establish a mechanism where the abstract semantics can actively perceive and retrieve specific geometric details. This module facilitates a deep interaction between the implicit emotion features and the explicit geometric priors. First, to ensure the retrieval is based purely on semantic intent rather than unstable intensity noise, we perform spherical semantic projection. Taking the expression branch as an example, given the implicit audio feature tensor $f^a_e \in \mathbb{R}^{B \times N \times m \times d}$, we apply $L_2$ normalization specifically along its feature dimension $d$. By projecting each $d$-dimensional feature vector onto a unit hypersphere, we formulate it as:
\begin{equation}
\hat{f}^a_e = \frac{f^a_e}{\|f^a_e\|_2 + \epsilon},
\end{equation}
where $\|\cdot\|_2$ denotes the $L_2$ norm computed over the $d$ dimension, and $\epsilon$ is a small constant for numerical stability. Geometrically, this operation effectively strips away the uninformative and uncontrollable magnitude variance while strictly preserving the correct semantic direction (i.e., the emotion category) encoded within the latent space. Subsequently, we employ the geometric querying mechanism to bridge the two modalities. The normalized vector $\hat{f}^a_e$ serves as the query to attend to the continuous geometric priors ${F}_\text{geo}$ (derived from D-GPG), which serve as both keys and values. Through this attention, the semantic feature adaptively \textbf{searches and aggregates} specific muscle activation patterns—such as the amplitude of jaw opening or eyebrow tension—matching the current emotion. This yields a geometry-aware context matrix $\boldsymbol{C}_{e}\in\mathbb{R}^{B\times N\times K\times d}$, encoding the precise structural instructions:
\begin{equation}
\boldsymbol{C}_{e} = \text{Attention}(\hat{f}^a_e {W}_Q,\, {F}_\text{geo} {W}_K,\, {F}_\text{geo} {W}_V).
\end{equation}

\begin{figure*}[t]
  \centering
  \includegraphics[width=1\textwidth, trim=0 40 51 0, clip]{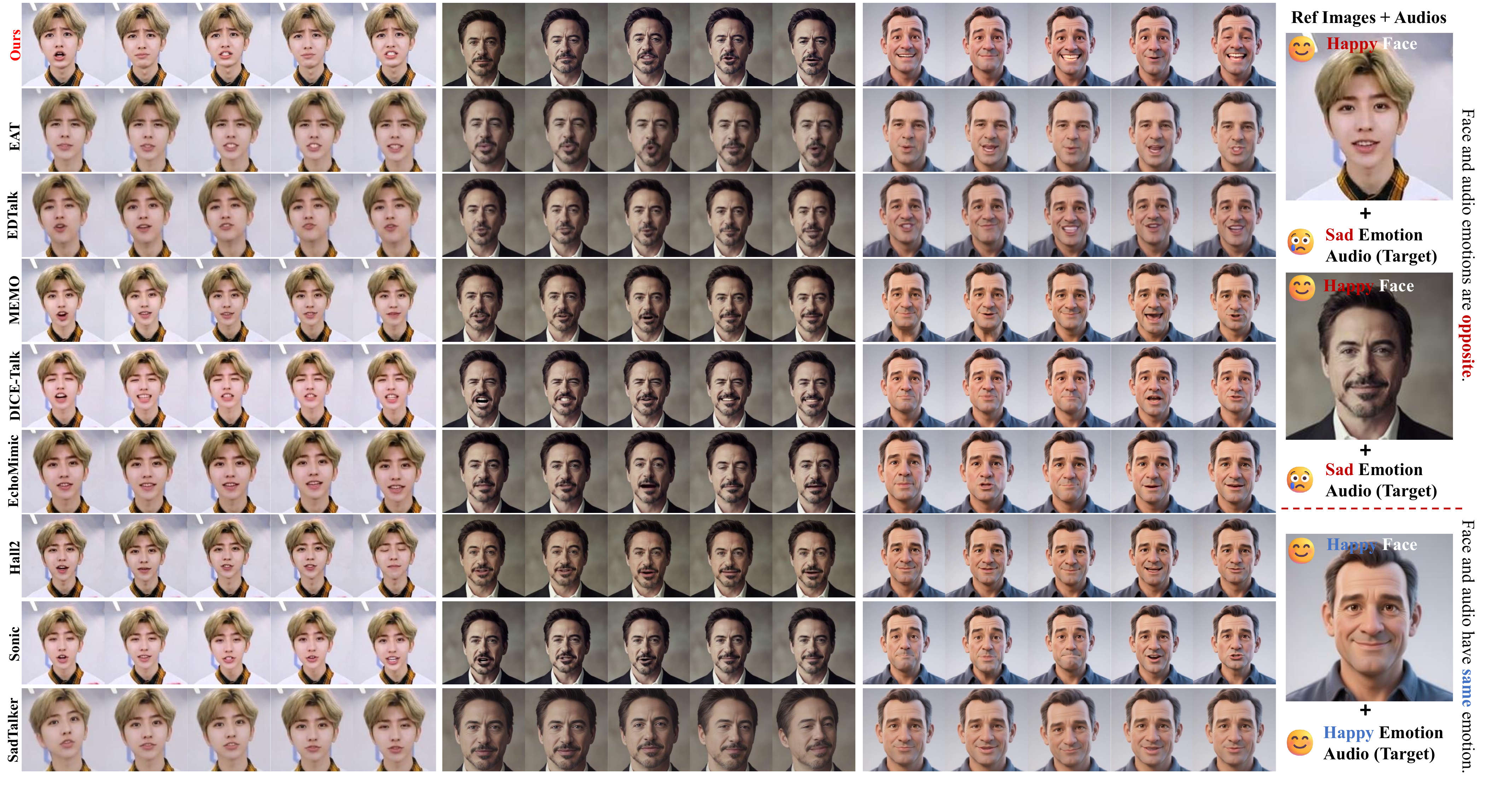}
  \caption{Qualitative comparisons with SOTA methods \textbf{on out of domain} data. Our method can obtain more accurate facial emotions, especially the emotions of the reference face image and the driving audio are opposite. Moreover, full video comparisons are available in the supplementary material to show our method's effectiveness in sync, naturalness, and stability.}
  \label{fig:main_com_exp}
\end{figure*}


\noindent\textbf{Geometric Affine Adapter (GAA)}. This stage translates the aggregated geometric context $\boldsymbol{C}_{e}$ into element-wise modulation tensors to \textbf{recalibrate} the feature magnitude. We introduce the GAA which projects $\boldsymbol{C}_{e}$ into scale $\boldsymbol{\gamma}$ and shift $\boldsymbol{\beta}$ tensors (with the same spatial-temporal dimensions as the emotional audio-exp features) to modulate the normalized semantics:
\begin{equation}
\hat{F}_e^a \!=\! (1 \!+\! \gamma) \!\odot\! \hat{f}^a_e \!+\! \beta, \, \text{where } (\gamma, \beta) \!=\! \text{Proj}(\mathbf{C}_{e}),
\end{equation}
where $\odot$ is Hadamard product. Crucially, we initialize the final projection layer of GAA to zero. This ensures training starts as an identity mapping ($\hat{F}^a_{e}=\hat{f}^a_{e}$), preserving the pre-learned semantic manifold. To prevent the modulation from rotating the feature vector (causing semantic drift), we impose a directional consistency loss: $\mathcal{L}_\text{dcl} = 1 - \text{sim}(\hat{F}^a_e, \hat{f}^a_e)$, where $\text{sim}(\cdot,\cdot)$ denotes cosine similarity. This constraint compels the network to encode structural intensity strictly into the \textbf{magnitude} ($\|\hat{F}^a_e\|$) while maintaining the original semantic direction. Finally, the recalibrated feature $\hat{F}^a_e$, now enriched with geometric magnitude, is injected into the denoising U-Net via cross attention. It serves as the source for keys and values to guide the generation:
\begin{equation}
\label{eq:gem}
\mathbf{o}^e_t \!=\! \text{Softmax}\!\left(\!\frac{(\mathbf{z}_t \!\cdot\! W_Q) \!\cdot\! (\hat{F}^a_e \!\cdot\! {W}_K)^\top}{\sqrt{d}}\!\right) \!\cdot\! (\hat{F}^a_e \!\cdot\! {W}_V),
\end{equation}
where $\mathbf{z}_t$ is the latent noise of the denoising U-Net, and $\mathbf{o}_t^e$ geometry-modulated expression conditional features injected into the U-Net intermediate layers. Similarly, the audio-exp representation $\mathbf{o}^l_t$ can be obtained by the same mechanism. To integrate these anatomically distinct signals, we employ a condition aggregation strategy: the two conditional features are aggregated via element-wise addition, projected, and injected into $\mathbf{z}_t$ via a residual connection. By explicitly coupling geometric intensity with the feature norm, the attention mechanism becomes intensity-aware. This modulation sharpens the attention distribution (via the inner product $Q\cdot K^\top$), transforming dispersed attention into focused activation on relevant facial regions, while simultaneously enhancing the signal injection strength (via the value). Consequently, during inference, this mechanism allows us to directly steer the denoising process by adjusting the values of blendshape coefficients under fixed emotional semantic constraints, realizing facial dynamics generation with continuous, geometrically defined emotion intensity.

\subsection{Training and Inference}
\label{sec:TRAINING}
\textbf{Training.} We implement a three-stage progressive training strategy: (1) We train the backbone to reconstruct target frames from reference images using arbitrary video clips. This stage optimizes the reference net and basic modules within the denoising U-Net, improving single-frame generation capability; (2) To establish robust articulation and isolate phonetic learning from affective variance, we inject the lip attention layers in the denoising U-Net, strictly restricting audio-visual interaction to the mouth region by using spatial lip masks, while Gaussian noise is applied to the global latent space. We freeze the reference net and this masked guidance compels the model to master precise lip-synchronization using large-scale neutral datasets. (3) We train the V-AEP and D-GPG modules and freeze their parameters to derive the coordinated emotional representations ($f^a_l, f^a_e$) and geometric priors ($F_\text{geo}$) as conditional inputs. Then, we train the final denoising U-Net with the GEM. To explicitly suppress reference emotion leakage, we employ a \textbf{conflict-aware training strategy} by sampling tuples with mismatched reference-audio emotions. The model is optimized via the standard diffusion objective:
\begin{equation}
    \mathcal{L}_\text{gem} \!=\! \mathbb{E}_{\mathbf{z}_t, \epsilon\sim\mathcal{N}(0,1), t} \big[ \|\epsilon \!-\! \epsilon_\theta(\mathbf{z}_t, t, f^a_l, f^a_e, F_\text{geo}) \|_2^2 \big],
\end{equation}
where $\epsilon_\theta$ denotes the denoising network, $\epsilon$ is the noise added to the latent $\mathbf{z}_t$ at timestep $t$. \textit{Details of training the lip attention layers and the conflict-aware training strategy are provided in the Appendix. }

\noindent\textbf{Inference.} The network takes a reference image and driving audio as inputs to generate emotionally and phonetically aligned videos. To enable long-duration synthesis, we employ a sliding-window generation strategy, utilizing the last $T$ frames of the preceding clip as motion priors to ensure temporal continuity. Crucially, the persistent identity guidance from the reference net effectively counteracts error accumulation, preventing identity drift throughout the generation process. Furthermore, we design an adaptive inter-frame \textbf{smoothing} strategy to effectively mitigate temporal jitter across both the subject and background (\textit{see the Appendix for details}).

\section{Experiments}
\label{sec:exp}
\subsection{Experiments Setup}
\textbf{Datasets.} We use HDTF (363 IDs, 15.8 hours) \cite{HDTF} as the neutral talking dataset. Following prior work \cite{vexpress,hallo}, we split HDTF into training and testing sets with a 9:1 ratio. We created a diverse dataset by aggregating data from the front face videos of MEAD (48 IDs, 31.2 hours) \cite{MEAD} and the speech part of RAVDESS (24 IDs, 3 hours) \cite{RAVDESS} as the emotional talking dataset with 8 emotion categories. We followed the standard identity-based train/test splits adopted in EAT \cite{EAT} to ensure subject-independent evaluation. We use all the training data to train our face reconstruction backbone, and use the HDTF training data to train the lip attention layers. The training data of MEAD and RAVDESS was used to train our V-AEP and GEM.  We also use a small part HDTF as neutral emotion data to train to prevents the model from overfitting to the specific studio conditions of the emotional dataset. 

\begin{table*}[t]
\centering
\caption{Quantitative comparisons on HDTF and MEAD (Front) + RAVDESS (Speech) datasets, with best in \textbf{bold}, second in \underline{underline}. Notably, our GemTalk outperforms all recent emotional talking face generation methods (\textbf{second raw}).}
\label{tab:main_com_exp}
\renewcommand{\arraystretch}{0.9} 
\resizebox{0.99\textwidth}{!}{
\begin{tabular}{c|ccccc|cccccc}
\toprule
\textsc{\textbf{Dataset}} & \multicolumn{5}{c|}{{\textbf{HDTF}}} & \multicolumn{6}{c}{{\textbf{MEAD (FRONT) + RAVDESS (Speech)}}} \\
\cmidrule(r){1-1} \cmidrule(lr){2-6} \cmidrule(l){7-12}

\textsc{\textbf{Method}} & 
FVD ($\downarrow$) & FID ($\downarrow$) & Sync-C ($\uparrow$) & Sync-D ($\downarrow$) & E-FID ($\downarrow$) & 
FVD ($\downarrow$) & FID ($\downarrow$) & Sync-C ($\uparrow$) & Sync-D ($\downarrow$) & E-FID ($\downarrow$) & $\text{Acc}_\text{emo}\%$ ($\uparrow$) \\ 
\midrule

\textsc{SadTalker} (2023) 
& 458.117 & 55.470 & 4.923 & 9.803 & 2.836 & 636.347 & 58.762 & 5.213 & 9.134 & 6.625 & 17.231 \\

\textsc{Hallo} (2024) 
& 219.380 & \underline{22.595} & \underline{7.835} & 7.862 & 2.634 & 376.630 & \underline{36.397} & 6.849 & \underline{8.522} & 3.815 & 17.817 \\

\textsc{Hallo2} (2024) 
& 225.734 & 25.466 & 7.398 & 7.988 & 2.462 & \underline{355.772} & 39.225 & 6.836 & 8.935 & 4.343 & 17.742 \\

\textsc{V-Express} (2024) 
& 617.420 & 43.831 & 4.066 & 10.836 & 3.122 & 728.437 & 79.752 & 4.489 & 10.367 & 6.835 & 16.943 \\

\textsc{AniPortrait} (2024) 
& 346.983 & 29.472 & 4.157 & 11.233 & 3.061 & 593.838 & 50.394 & 3.587 & 12.539 & 7.362 & 17.539 \\

\textsc{EchoMimic} (2025) 
& 275.640 & 24.513 & 4.838 & 10.356 & 1.568 & 462.410 & 56.352 & 5.641 & 10.133 & 5.983 & 18.432 \\

\textsc{Sonic} (2025)
& \textbf{175.562} & \textbf{21.351} & \textbf{7.993} & \underline{7.531} & \underline{1.437} & 363.541 & 38.658 & \textbf{7.493} & 8.697 & 3.886 & 19.216 \\
\midrule

\textsc{DreamTalk} (2023) 
& 393.132 & 43.953 & 6.507 & 8.261 & 2.817 & 560.336 & 61.877 & 5.739 & 9.261 & 6.742 & 51.343 \\

\textsc{EAT} (2023) 
& 593.127 & 58.332 & 6.340 & 8.029 & 3.177 & 626.029 & 72.883 & 6.252 & 8.906 & 6.866 & 52.302 \\

\textsc{EDTalk} (2024)
& 496.226 & 50.984 & 6.639 & 8.106 & 2.887 & 585.611 & 53.293 & 5.827 & 9.452 & 6.687 & \underline{55.926} \\

\textsc{MEMO} (2025) 
& 233.946 & 32.721 & 7.659 & 8.370 & 2.835 & 427.754 & 42.910 & 6.419 & 9.651 & 4.179 & 43.847 \\

\textsc{DICE-Talk} (2025)
& 239.750 & 25.173 & 7.526 & 8.192 & 1.529 & 363.256 & 37.429 & 6.805 & 9.024 & \underline{3.738} & 49.439 \\
\midrule
\textsc{GemTalk}
& \underline{216.138} & 23.136 & 7.819 & \textbf{7.449} & \textbf{1.386} & \textbf{334.937} & \textbf{31.625} & \underline{7.027} & \textbf{8.283} & \textbf{2.392} & \textbf{59.258} \\

\bottomrule
\end{tabular}}
\end{table*}

\noindent\textbf{Evaluation Metrics.} The Fréchet Inception Distance (FID) \cite{FID} and Fréchet Video Distance (FVD) \cite{FVD} measure measure the similarity between generated images and real data. The Synchronization-C (Sync-C) and Synchronization-D (Sync-D) \cite{sync} assess the lip synchronization quality, with higher Sync-C and lower Sync-D scores indicating better alignment with the driven speech signal. Expression-FID (E-FID) \cite{emo} evaluates the expression divergence between the synthesized videos and ground-truth videos. To assess the emotional accuracy $\text{Acc}_\text{emo}$ of the generated facial emotions, we use Emotion-Fan \cite{emo-score} to calculate the average emotional accuracy.

\begin{figure*}[t]
  \centering
  \includegraphics[width=1\textwidth, trim=0 0 0 0, clip]{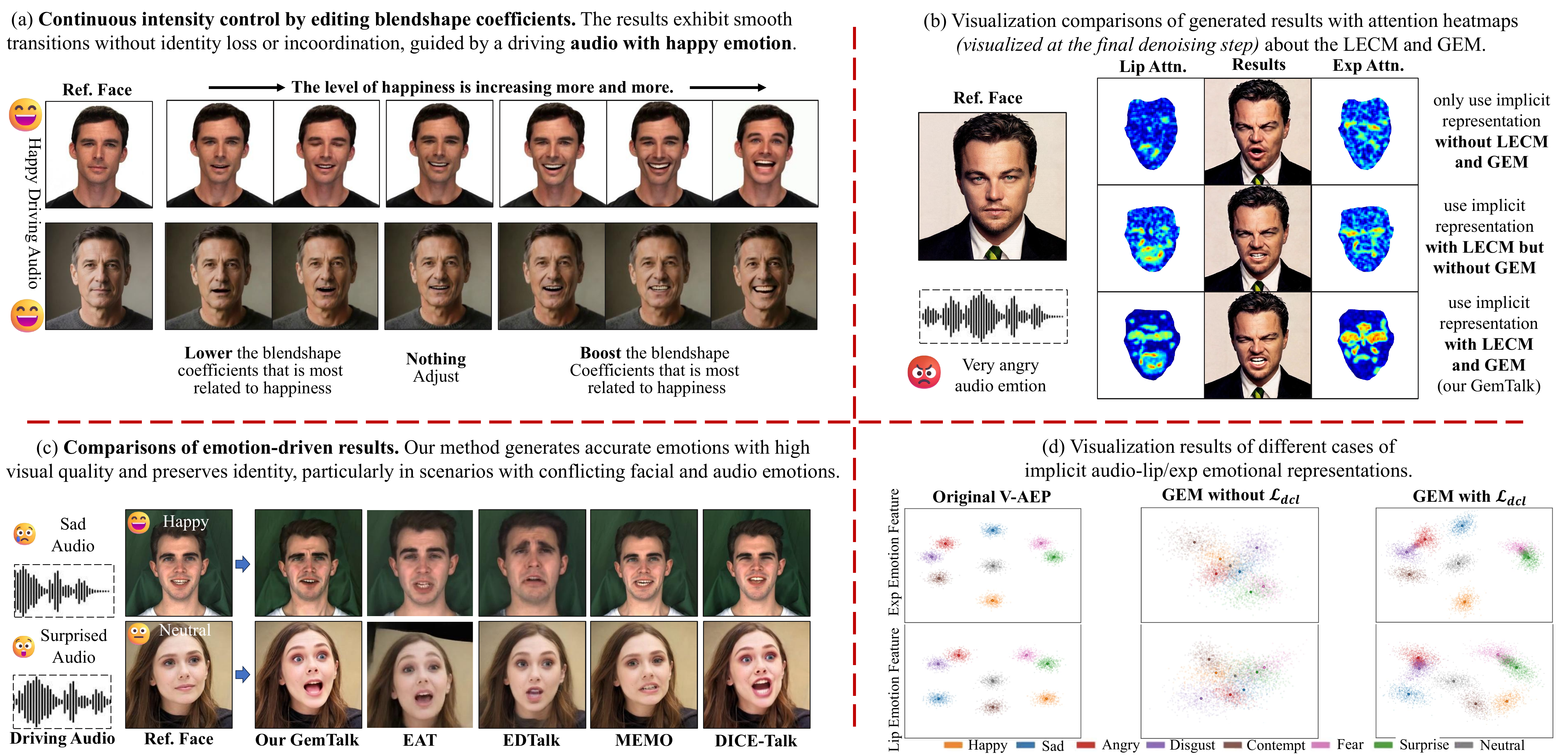}
  \caption{Visual results of ablation studies. (a) Visual comparisons of continuous intensity control by editing blendshape coefficients. (b) Visualization comparisons of generated results with attention heatmaps about the LECM and GEM. (c) Visual comparisons of four emotional talking face generation methods. (d) Visualization results of different cases of implicit audio-lip/exp emotional representations.}
  \label{fig:ablation}
\end{figure*}

\noindent\textbf{Implementation Details.} Our experiments encompassing both training and inference are conducted on 4 NVIDIA A6000 GPUs. Videos are sampled at 25 FPS with audio at 16 kHz, where each training clip consists of $N=16$ frames. For the first two stages, D-GPG and GEM training, frames are resized to $512\times512$. Each stage is trained for 30,000 steps with a batch size of 2, using the AdamW optimizer with a fixed learning rate of $1\times10^{-5}$. During the GEM training stage, to ensure video continuity, the model produces 14 video frames with the motion module's latents concatenated with the first 2 frames. The motion module is initialized with pre-trained weights from AnimateDiff. Furthermore, to enhance video generation robustness and enable classifier-free guidance, the reference image, guidance audio, and motion frames are randomly dropped with a 0.05 probability during training. Conversely, V-AEP is trained on $256\times256$ images for 300 epochs with a batch size of 32, also using the AdamW optimizer with a learning rate of $1\times10^{-4}$.

\subsection{Main Results}
To demonstrate the effectiveness of our purely audio-driven framework, we primarily compare our method with recent state-of-the-art purely audio-driven talking face generation works. 

\noindent\textbf{Quantitative Comparisons.} In Table \ref{tab:main_com_exp}, Our GemTalk outperforms other methods across most metrics on the MEAD + RAVDESS dataset, an aggregated emotional dataset, ranking second only in FID. Notably, it surpasses the second-best by $3.33\%$ in $\text{Acc}_\text{emo}$ and reduces FVD by $20.84$, reflecting our method synthesize accurate emotions and high quality videos. Furthermore, On the non-emotional HDTF dataset, GemTalk maintains superior performance in Sync-D and E-FID, while securing the second-best FVD among non-emotional methods. These results validate that our method (GEM) effectively achieves diversity and accuracy in emotional expression without compromising lip-sync precision and visual quality for neutral emotion data. 

\noindent\textbf{Qualitative Comparisons.} Figure \ref{fig:main_com_exp} provides visualization
comparisons on out of domain data. Analysis results show that our method demonstrates the richest and most superior emotional expressiveness than any other methods, while other methods either fail to manifest the intended emotion or result in unnatural facial dynamics. Compared to recent emotion-agnostic methods, our approach maintains high identity preservation and image quality while offering a diverse range of emotional expressions. Compared to the emotional models, our method generates significantly more natural and temporally coherent expressions, while other models often suffer from rigid expressions or excessive facial distortions, which may lead to identity inconsistency. We show more emotion generation results in Figure \ref{fig:ablation}.

\noindent\textbf{User Study.} We conducted a user study to assess comparative methods in 4 dimensions: emotional expression, identity consistency, lip sync, and video smoothness, with 50 participants to rated score (1 to 5) to 100 randomly selected videos. Here, we assessed our adaptive sliding-window smoothing strategy, because it will randomly  reduce frames, making it unsuitable for comparison against real videos. As shown in Table \ref{tab:user_study}, our method outperforms others in emotional expression. 

\subsection{Ablation Study}

To validate each component, we conduct a comprehensive ablation study (Table \ref{tab:ablation}). Progressively integrating the implicit lip/expression representations, LECM, and GEM, our full model GemTalk achieves the best performance. Figure \ref{fig:ablation}(c) further shows our superiority over emotion-driven methods (e.g., EAT, DICE-Talk), synthesizing more accurate and high-fidelity expressions with strong robustness even under conflicting audio-visual emotional cues.

\begin{figure}[t]
  \centering
  \vspace{-1.2pt} 
  \includegraphics[width=1 \linewidth, trim=0 0 0 0, clip]{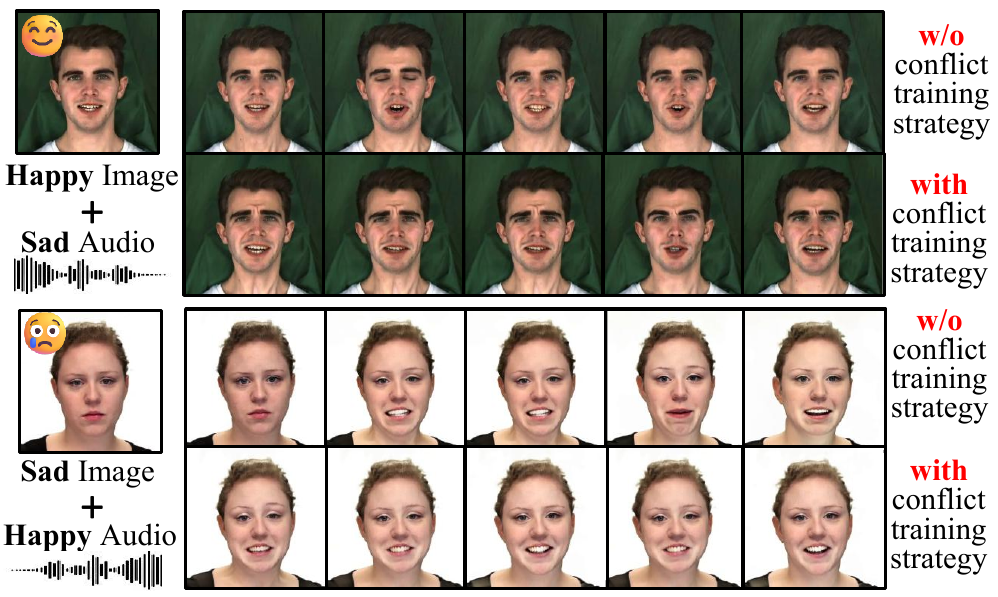}
  \caption{Ablation study of conflict-aware training strategy.}
   \label{conflict_ablation}
  \vspace{-3.2pt} 
\end{figure}

\begin{table}[t]
\centering
\caption{{Quantitative ablation study on MEAD+RAVDESS datasets.} We progressively integrate components into M1 (emotion-agnostic backbone) to validate their effectiveness.}
\label{tab:ablation}
\resizebox{0.98\columnwidth}{!}{
\begin{tabular}{l|cccc}
\toprule
\textbf{Method} & {FVD} ($\downarrow$) & {FID} ($\uparrow$) & {Sync-C} ($\downarrow$) & {$\text{Acc}_\text{emo}\%$} ($\uparrow$) \\ 
\midrule
M1: Backbone 
& 395.081 & 43.397 & \textbf{7.125} & 17.125 \\

M2: + Implicit exp 
& 367.412 & 38.436 & 7.108 & 49.259 \\

M3: + Implicit lip 
& 362.733 & 38.127 & 6.897 & 51.932 \\

M4: + LECM 
& 358.129 & 35.528 & 6.960 & 53.618 \\
\midrule
$\text{M5}^{\textbf{-}}$: + GEM (w/o $\mathcal{L}_\text{dcl}$) 
& 355.386 & 36.855 & 7.023 & 32.851 \\

$\text{M5}^{\textbf{+}}$: + GEM (with $\mathcal{L}_\text{dcl}$) (\textbf{full}) 
& \textbf{334.937} & \textbf{31.625} & {7.027} & \textbf{59.258} \\ 
\bottomrule
\end{tabular}
}
\end{table}
\begin{table}[t]
\centering
\renewcommand{\arraystretch}{0.9} 
\setlength{\tabcolsep}{3pt}
\caption{User study results on on out of domain data.}
\label{tab:user_study}
\resizebox{0.97\columnwidth}{!}{%
    \begin{tabular}{lcccc}
    \toprule
    \textsc{\textbf{Method}} & {Emotion} ($\uparrow$) & {Lip Sync} ($\uparrow$) & {ID Consistency} ($\uparrow$) & {Smoothness} ($\uparrow$) \\
    \midrule
    \textsc{EDTalk}   & 3.106 & 2.761 & 2.787 & 1.612 \\
    \textsc{DICE-Talk} & 3.583 & 2.665 & 3.219 & 2.667 \\
    \textsc{Hallo2}   & 2.166 & 3.702 & 3.912 & 3.485 \\
    \textsc{Sonic}    & 1.912 & \underline{4.269} & \underline{4.135} & \textbf{3.896} \\
    \midrule
    {\textsc{GemTalk}} & \textbf{4.223} & \textbf{4.296} & \textbf{4.187} & \text{3.691} \\
    
    {\textsc{GemTalk}+Smooth} 
    & \underline{4.193} & {4.232} & {4.081} & \underline{3.839} \\
    \bottomrule
    \end{tabular}%
}
\end{table}

\noindent\textbf{Fine-grained Controllability.} The GEM module endows our framework with geometric awareness, enabling continuous intensity control as a demonstration of robust controllability. We modulate the top-10 blendshape coefficients most correlated with the target emotion, selected based on their mean magnitude and high-activation frequency (0.5-1.0). To maintain muscle synergy, we employ proportional scaling rather than additive offsets, coupled with a safety boundary constraint where coefficients are clamped within dataset statistical extrema extended by a margin of 0.15, thereby preventing facial imbalance that would otherwise lead to control failure. As visually corroborated in Figure \ref{fig:ablation}(a) and Figure \ref{fig:intro_pic}, this strategy achieves seamless continuous control. Facial dynamics evolve naturally from subtle expressions to intense ones (e.g., a grin expanding into a broad laugh). Crucially, even under substantial coefficient scaling, the generated faces maintain strict anatomical coordination and identity consistency, validating that our geometric modulation achieves coordinated control over facial dynamics without distorting the facial structure.

\noindent\textbf{Effect of Progressive Refinement Strategy.} As shown in Table \ref{tab:ablation}, integrating the expression and lip branches (M2, M3) significantly boosts $\text{Acc}_\text{emo}\%$ over the backbone (M1) without compromising lip-sync fidelity. The attention heatmaps in Figure \ref{fig:ablation}(b) further reveal the anatomical impact: without structural guidance (M3), attention is dispersed with weak responses and background noise; with LECM (M4), the Lip attention begins to perceive upper facial contours, forming spatial synergy between lip and expression regions; with GEM ($\text{M5}^{\textbf{+}}$), activations become intense and focused, precisely targeting the corrugated brows and tensioned jaw for the very angry input. This confirms that GEM anchors emotion to specific geometric structures, yielding the most vivid generation.

\noindent\textbf{Effect of Directional Consistency Loss.} Figure \ref{fig:ablation}(d) visualizes the audio representations to analyze the effect of $\mathcal{L}_{dcl}$. The original V-AEP embeddings (M4) form compact clusters with clear inter-class margins, showing that V-AEP learns a discriminative emotional manifold. With $\mathcal{L}_{dcl}$ ($\text{M5}^{\textbf{+}}$), the modulated embeddings $\mathbf{o}_t$ remain largely separable, indicating that GEM injects geometry-guided modulation while preserving V-AEP's emotion semantics. In contrast, removing $\mathcal{L}_{dcl}$ ($\text{M5}^{\textbf{-}}$) collapses this structure, making different emotions indistinguishable.

\noindent\textbf{Conflict-aware Training Strategy.} To prevent \emph{reference emotion leakage}—where the model copies the static emotion of the reference image instead of following the driving audio—we introduce a conflict-aware training strategy. As shown in Figure \ref{conflict_ablation}, we test models under conflicting conditions (top: sad audio + happy reference; bottom: happy audio + sad reference). Without conflict training, emotion leakage occurs (e.g., sad eyes with a smile), causing uncoordinated facial dynamics. Our strategy ensures the generated emotion is strictly controlled by the audio.

\section{Conclusion} 
Motivated by the tension between controllability and high fidelity in audio-driven emotional talking-face generation, we propose \textsc{GemTalk}. Building upon a profound understanding of attention computation, it integrates vision-guided implicit audio emotion projection with diffusion-based explicit geometric priors. Specifically, our geometry-guided emotion modulation enhances implicit representations without compromising fine-grained semantics, enabling continuous, interpretable control while preserving realistic appearance and stable lip-expression coordination. Experiments show that \textsc{GemTalk} improves perceptual quality, lip-sync consistency, and emotion accuracy, supporting smooth fine-grained emotion editing via geometric priors. Future work will explore robustness in in-the-wild scenarios (e.g., large pose, occlusion, noisy audio) and extend geometry-guided control to richer head/upper-body expressions.

\begin{acks}
This work is funded by National Natural Science Foundation of China (62576305) and Ningbo Youth Science and Technology Innovation Talent Project (2025QL052).
\end{acks}



\bibliographystyle{ACM-Reference-Format}
\bibliography{sample-base}


\appendix
\twocolumn[{%
  \centering
  {\Huge\sffamily\bfseries Geometry-guided Emotion Modulation for Controllable and Photorealistic Emotional Talking Face Generatio\par}
  \vspace{0.6em}
  {\Large\sffamily Supplementary Material\par}
  \vspace{1.5em}
}]

\section{Blendshape Coefficients }
\label{app_sec_bc}
To provide the implicit representations with physically interpretable structural guidance, we utilize the standard Apple ARKit blendshape definition. This parametrization decomposes facial dynamics into $52$ distinct muscle activation coefficients, where each coefficient is a scalar value in the range $[0, 1]$ quantifying muscle intensity. Crucially, this definition naturally aligns with our framework's emotional modeling strategy by explicitly covering both lip-related (e.g., jaw, mouth) and expression-related (e.g., eyes, eyebrows, cheeks) regions. Since these coefficients are derived from real facial captures, they inherently preserve the anatomical synergy and coordination between lip movements and facial expressions during normal speech. The complete list of the $52$ blendshape bases is provided in Table \ref{app_tab:full_bs_list}, where valid dimensions are in black and discarded ones are in \textcolor{red}{red}.

\begin{figure*}[h]
    \centering
    \includegraphics[width=0.98\linewidth]{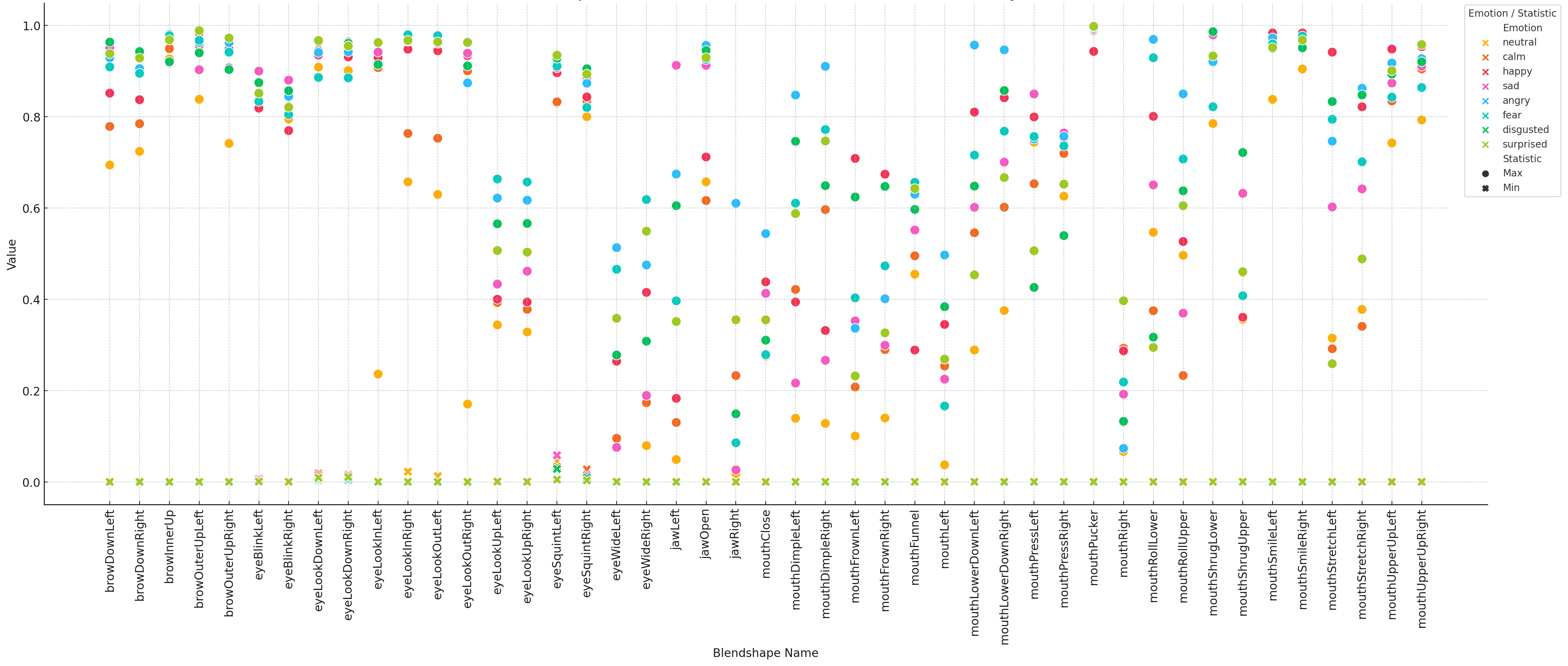}
    \caption{The maximum and minimum value distribution of the remaining $K=45$ blendshape coefficients. The high dynamic range in these dimensions indicates their significance in modeling lip synchronization and facial expressions.}
    \label{app_fig:bs_distribution}
\end{figure*}

However, not all blendshape dimensions are relevant for audio-driven talking face generation. Certain coefficients are rarely activated during normal speech. To identify the effective dimensions, we performed a comprehensive statistical analysis on the aggregated MEAD (Front) and RAVDESS (Speech) datasets. This analysis covered 8 emotion categories (Angry, Disgust, Contempt, Fear, Happy, Sad, Surprise, and Neutral), explicitly excluding ``Calm'' to ensure rigorous emotional intensity standards. We calculated the \textbf{maximum activation value} for each dimension across all emotional clips. We observed that 7 dimensions exhibited negligible activations, with maximum values consistently remaining below the threshold of $0.05$. Specifically, the dimensions marked in \textcolor{red}{red} in Table \ref{app_tab:full_bs_list}—including \texttt{cheekPuff} (max $\approx 0.0037$), \texttt{jawForward} (max $\approx 0.0105$), and others like \texttt{noseSneer} and \texttt{cheekSquint} (max $< 0.0001$)—are statistically insignificant compared to active speech muscles. Consequently, we discard these 7 sparse dimensions, resulting in a compact and effective geometric space of $K=45$ dimensions. We visualize the distribution of maximum and minimum activation values for the remaining 45 coefficients in Figure \ref{app_fig:bs_distribution}. 
\begin{figure*}[h]
    \centering
    \begin{subfigure}[b]{0.20\linewidth}
        \centering
        \includegraphics[width=\linewidth]{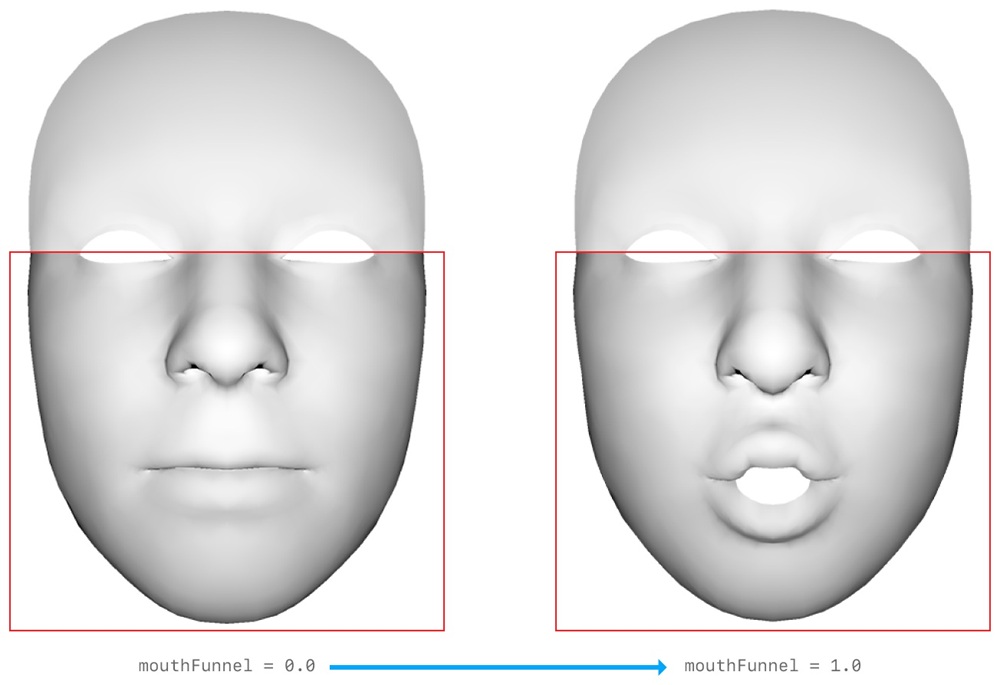}
        \caption{\texttt{mouthFunnel}}
        \label{fig:bs_lip1}
    \end{subfigure}
    \hfill
    \begin{subfigure}[b]{0.20\linewidth}
        \centering
        \includegraphics[width=\linewidth]{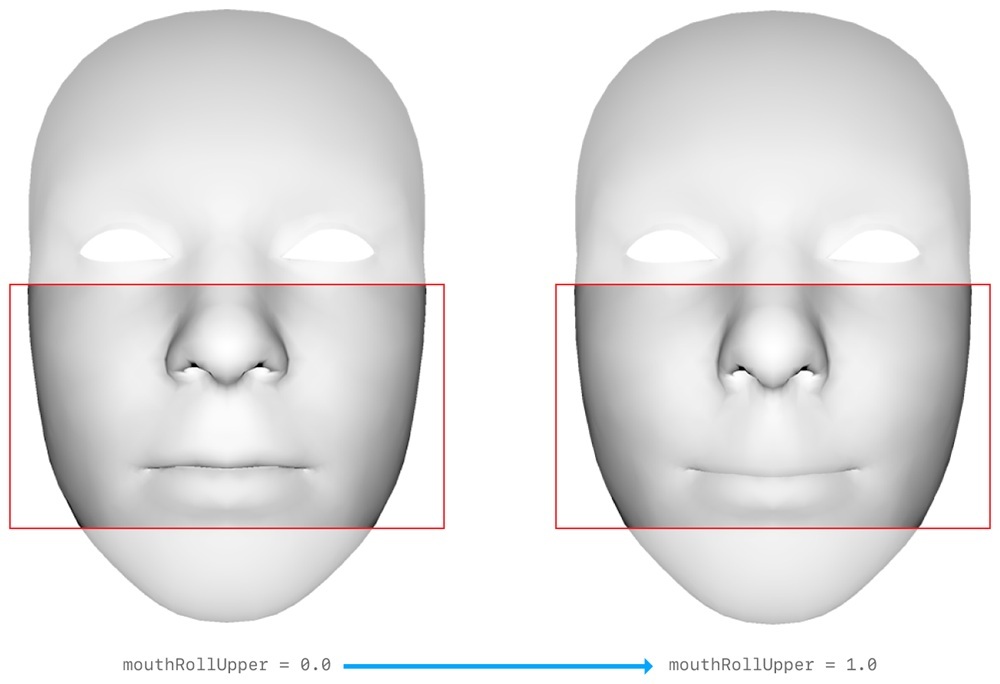}
        \caption{\texttt{mouthRollUpper}}
        \label{fig:bs_lip2}
    \end{subfigure}
    \hfill
    \begin{subfigure}[b]{0.20\linewidth}
        \centering
        \includegraphics[width=\linewidth]{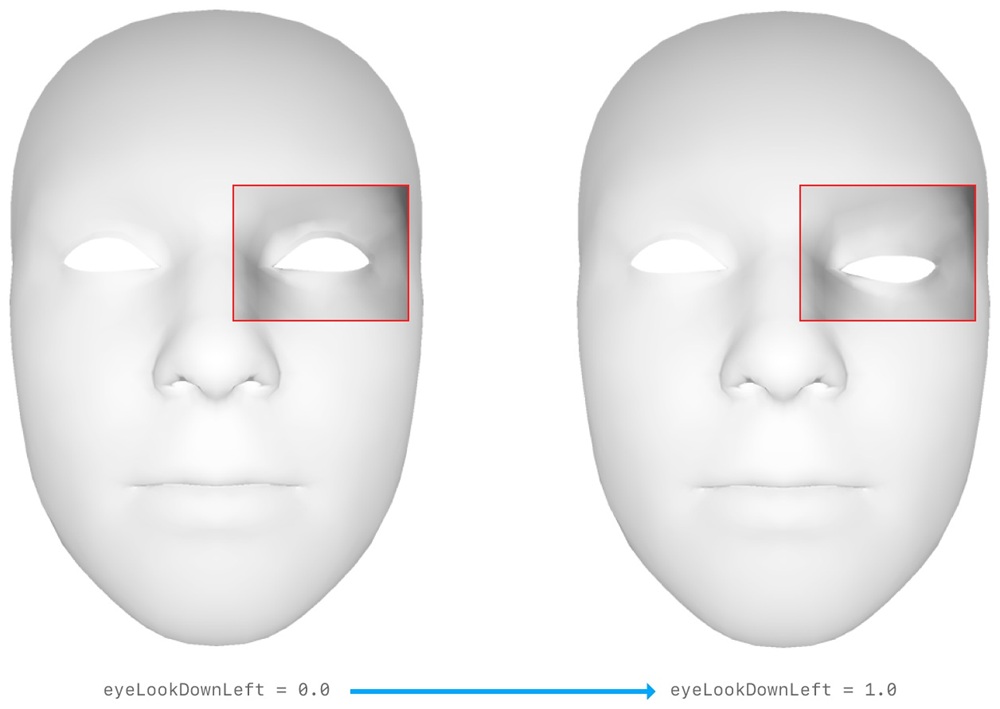}
        \caption{\texttt{eyeLookDownLeft}}
        \label{fig:bs_exp1}
    \end{subfigure}
    \hfill
    \begin{subfigure}[b]{0.20\linewidth}
        \centering
        \includegraphics[width=\linewidth]{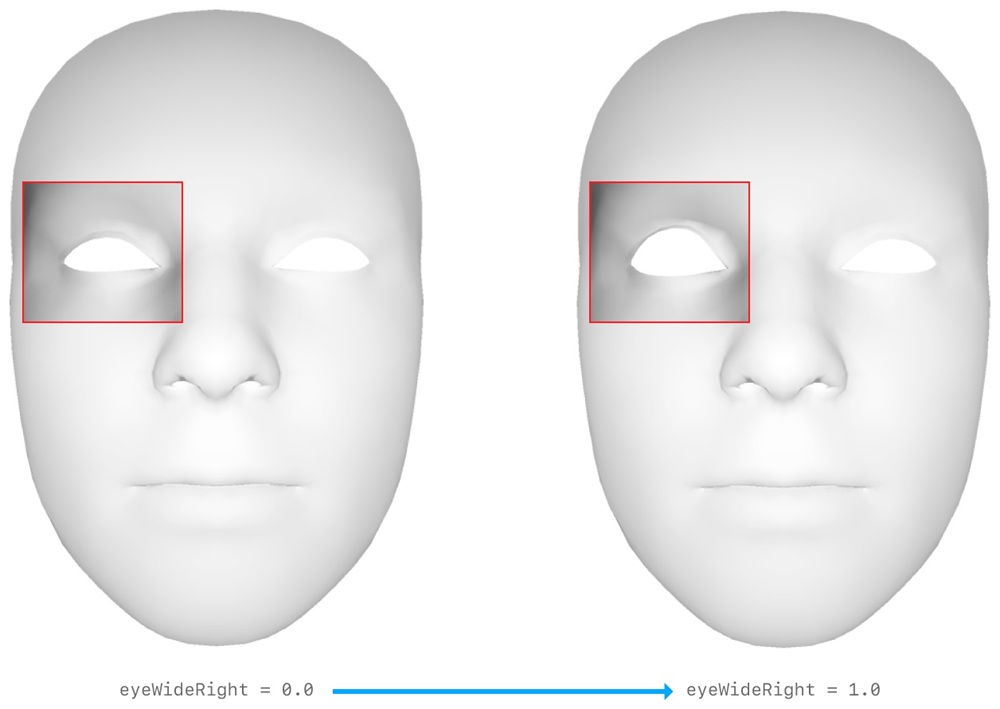}
        \caption{\texttt{eyeWideRight}}
        \label{fig:bs_exp2}
    \end{subfigure}
    
    \caption{Visualization of four distinct ARKit blendshape bases\protect\footnotemark (image source from the open-source Apple ARKit Documentation). (a) and (b) belong to the \textbf{Lip region}, controlling mouth shape and rolling dynamics. (c) and (d) belong to the \textbf{Expression region}, controlling eye gaze and openness. These bases act as physically defined geometric priors in our framework.}
    \label{app_fig:bs_vis}
\end{figure*}
\footnotetext{Image source: Apple ARKit Documentation.}

To vividly illustrate the physical semantics of these scalar coefficients, we visualize four specific blendshape bases in Figure \ref{app_fig:bs_vis}. As shown, \texttt{mouthFunnel} and \texttt{mouthRollUpper} ((a) and (b)) explicitly control the fine-grained deformation of the lip and mouth region. Conversely, \texttt{eyeLookDownLeft} and \texttt{eyeWideRight} ((c) and (d)) govern the upper facial movements, providing precise structural details for eye-related emotional expression.

\begin{table*}[h]
\centering
\caption{The complete list of 52 Apple ARKit Blendshape dimensions. The 7 dimensions marked in \textcolor{red}{\textbf{red}} exhibit negligible activation (max value $< 0.05$) on the MEAD+RAVDESS dataset and are discarded in our framework.}
\label{app_tab:full_bs_list}
\resizebox{0.8\linewidth}{!}{
\begin{tabular}{llll}
\toprule
0: \textcolor{red}{\texttt{\_neutral}} & 13: \texttt{eyeLookInLeft} & 26: \texttt{jawRight} & 39: \texttt{mouthRight} \\
1: \texttt{browDownLeft} & 14: \texttt{eyeLookInRight} & 27: \texttt{mouthClose} & 40: \texttt{mouthRollLower} \\
2: \texttt{browDownRight} & 15: \texttt{eyeLookOutLeft} & 28: \texttt{mouthDimpleLeft} & 41: \texttt{mouthRollUpper} \\
3: \texttt{browInnerUp} & 16: \texttt{eyeLookOutRight} & 29: \texttt{mouthDimpleRight} & 42: \texttt{mouthShrugLower} \\
4: \texttt{browOuterUpLeft} & 17: \texttt{eyeLookUpLeft} & 30: \texttt{mouthFrownLeft} & 43: \texttt{mouthShrugUpper} \\
5: \texttt{browOuterUpRight} & 18: \texttt{eyeLookUpRight} & 31: \texttt{mouthFrownRight} & 44: \texttt{mouthSmileLeft} \\
6: \textcolor{red}{\texttt{cheekPuff}} & 19: \texttt{eyeSquintLeft} & 32: \texttt{mouthFunnel} & 45: \texttt{mouthSmileRight} \\
7: \textcolor{red}{\texttt{cheekSquintLeft}} & 20: \texttt{eyeSquintRight} & 33: \texttt{mouthLeft} & 46: \texttt{mouthStretchLeft} \\
8: \textcolor{red}{\texttt{cheekSquintRight}} & 21: \texttt{eyeWideLeft} & 34: \texttt{mouthLowerDownLeft} & 47: \texttt{mouthStretchRight} \\
9: \texttt{eyeBlinkLeft} & 22: \texttt{eyeWideRight} & 35: \texttt{mouthLowerDownRight} & 48: \texttt{mouthUpperUpLeft} \\
10: \texttt{eyeBlinkRight} & 23: \textcolor{red}{\texttt{jawForward}} & 36: \texttt{mouthPressLeft} & 49: \texttt{mouthUpperUpRight} \\
11: \texttt{eyeLookDownLeft} & 24: \texttt{jawLeft} & 37: \texttt{mouthPressRight} & 50: \textcolor{red}{\texttt{noseSneerLeft}} \\
12: \texttt{eyeLookDownRight} & 25: \texttt{jawOpen} & 38: \texttt{mouthPucker} & 51: \textcolor{red}{\texttt{noseSneerRight}} \\
\bottomrule
\end{tabular}
}
\end{table*}


\section{Loss Functions Details of The V-AEP}
\label{app_sec_V-AEP}
The core objective of V-AEP is to bridge the inherent audio-visual modality gap, extracting emotion-related features from audio that are specifically optimized for visual generation tasks. To achieve this, we first establish a robust visual emotion space via contrastive learning to isolate emotional semantics from identity and noise. Subsequently, rather than performing strict one-to-one alignment which limits diversity, we employ a {stochastic multi-target alignment strategy}. This approach maps audio features to the average representation of randomly selected visual anchors sharing the same emotion. By avoiding rigid alignment, this strategy ensures the learned audio representations ($f_l^a, f_e^a$) capture generalized emotional dynamics suitable for driving diverse facial expressions.

\noindent\textbf{Visual Emotional Representations Encoding.} To construct a visual space that purely encodes emotional semantics while discarding identity attributes and environmental noise, we employ a robust contrastive learning strategy enhanced by aggressive data augmentation and hard negative mining. Before feature extraction, we apply a specific set of data augmentations to the input video frames. Since identity and lighting information largely reside in texture and color, we aim to force the visual encoder to rely solely on geometric structural deformations (e.g., muscle movements). Specifically, we apply \texttt{RandomHorizontalFlip}, \texttt{RandomRotation}, and severe color distortions using \texttt{ColorJitter}, \texttt{RandomGrayscale}, and \texttt{GaussianBlur}. By destructing low-level visual cues, the network is compelled to learn features invariant to the subject's appearance. Following feature extraction, we utilize Supervised Contrastive Learning with an {explicit hard negative mining strategy}. Taking the expression branch as an example, for each anchor sample $i$, we define the positive set $P(i)$ as the collection of indices for all other samples in the batch that share the same emotion label as $i$. Simultaneously, we identify the hard negative set $\mathcal{N}_{hard}(i)$, containing the top-$K$ ($K=16$) samples with different labels that exhibit the highest dot product similarity with the anchor. The loss is computed over the union set $\mathcal{S}(i) = P(i) \cup \mathcal{N}_{hard}(i)$:

\begin{equation}
\mathcal{L}_\text{exp}^\text{v} = \sum_{i=1}^{N} \frac{-1}{|P(i)|} \sum_{p \in P(i)} \log \frac{\exp\left( f_e^v(i) \cdot f_e^v(p) / \tau \right)}{\sum_{k \in \mathcal{S}(i)} \exp\left( f_e^v(i) \cdot f_e^v(k) / \tau \right)},
\end{equation}

where $\tau$ is the temperature parameter. The lip branch follows the same procedure to compute $\mathcal{L}_\text{lip}^\text{v}$. The total visual objective is $\mathcal{L}_\text{visual} = \lambda_1 \mathcal{L}_\text{exp}^\text{v} + \lambda_2 \mathcal{L}_\text{lip}^\text{v}$, where we set $\lambda_1=0.6$ and $\lambda_2=0.4$.

\noindent\textbf{Audio Emotional Representations Projection.} Leveraging the robust, identity-invariant visual emotion space mentioned above, we train the audio projectors to map acoustic signals into this visual manifold. The primary challenge here is the inherent \textit{one-to-many} mapping: a single audio emotion can correspond to diverse facial expressions. Enforcing a strict one-to-one alignment forces the model to overfit to identity-specific details. To address this, we propose a {Stochastic Multi-target Alignment Strategy}. Instead of aligning with a single target, we align the projected audio feature with the centroid of multiple randomly selected visual anchors from the current batch. Specifically, for an input audio sample $i$ with emotion label $y_i$, we identify all visual samples in the batch sharing the same label. We randomly select a subset of these candidates, denoted as $\mathcal{V}_i$, with the size constrained to a maximum of $k=4$ (i.e., $1 \le |\mathcal{V}_i| \le 4$). If fewer than $k$ candidates are available, we use all of them; if no matching visual samples exist, the loss for this sample is skipped. The audio feature is then aligned to the mean vector of this set. Taking the expression branch as an example:

\begin{equation}
\begin{split}
&\mathcal{L}_\text{a}^\text{exp} = \sum_{i \in \text{valid}} \left( 1 - \frac{f_e^a(i) \cdot \bar{f}_e^v(i)}{\|f_e^a(i)\| \|\bar{f}_e^v(i)\|} \right), \\
&\quad \text{where} \quad \bar{f}_e^v(i) = \frac{1}{|\mathcal{V}_i|} \sum_{j \in \mathcal{V}_i} f_e^v(j).
\end{split}
\end{equation}

Here, $f_e^a(i)$ is the projected audio expression feature, and $\bar{f}_e^v(i)$ is the dynamic visual prototype. This strategy ensures the audio representations capture the generalized emotional essence. The same strategy is applied to the lip branch to compute $\mathcal{L}_\text{a}^\text{lip}$. The total audio projection loss is $\mathcal{L}_\text{audio} = \mathcal{L}_\text{a}^\text{exp} + \mathcal{L}_\text{a}^\text{lip}$.

\begin{figure*}[h]
    \centering
    \begin{subfigure}[b]{0.41\linewidth}
        \centering
        \includegraphics[width=\linewidth]{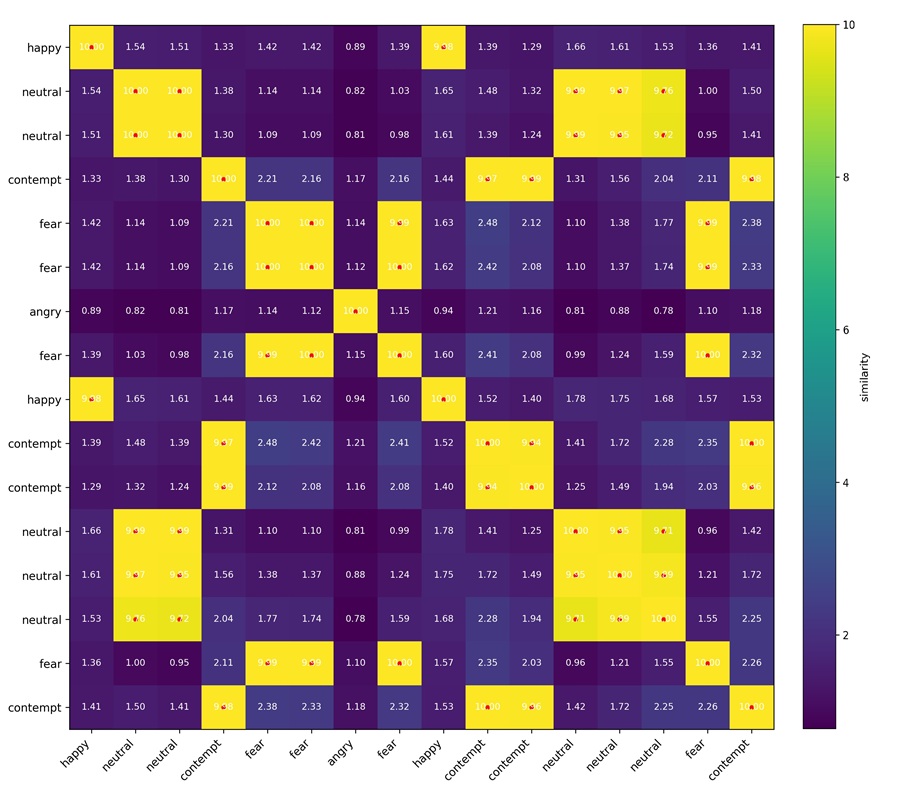}
        \caption{Audio-Lip Features ($f_l^a$)}
        \label{fig:vaep_lip_sim}
    \end{subfigure}
    \hspace{0.2cm} 
    \begin{subfigure}[b]{0.41\linewidth}
        \centering
        \includegraphics[width=\linewidth]{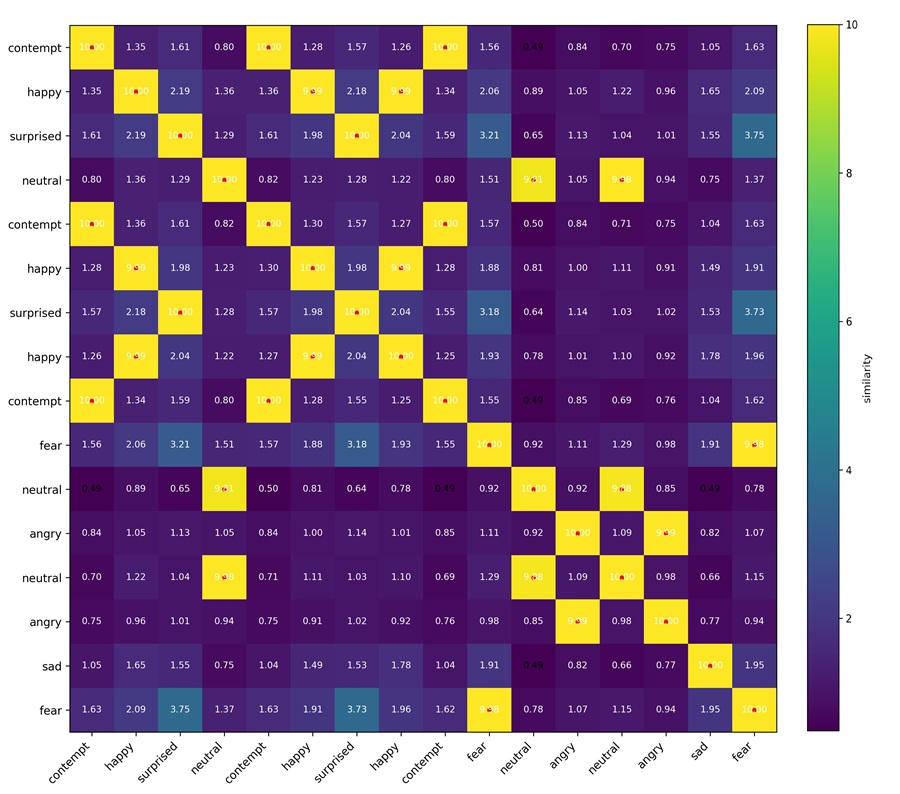}
        \caption{Audio-Exp Features ($f_e^a$)}
        \label{fig:vaep_exp_sim}
    \end{subfigure}
    
    \caption{Cosine similarity matrices of the projected audio features for a random batch of 16 samples. (a) shows the similarity in the lip emotion space, and (b) shows the expression emotion space.}
    \label{app_fig:vaep_similarity}
\end{figure*}

To intuitively verify the effectiveness of our Audio Emotional Representations Projection, we visualize the cosine similarity matrices of the learned audio-lip ($f_l^a$) and audio-expression ($f_e^a$) features in Figure \ref{app_fig:vaep_similarity}. We randomly selected 16 audio samples covering various emotion categories from the test set. In the heatmaps, the grid cells marked with \textcolor{red}{red dots} indicate positive pairs (i.e., pairs sharing the same ground-truth emotion label), while the remaining cells represent negative pairs. The positive pairs consistently exhibit high similarity scores (indicated by bright yellow colors), whereas negative pairs show significantly lower similarity (dark blue or purple colors). This clear separation demonstrates that, despite the stochastic nature of our alignment strategy, the V-AEP module successfully extracts robust and discriminative emotional semantics from audio, effectively aligning them with the visual emotion manifold.

\section{Details of Training Strategy and Inference Smoothing}
\label{app_sec_tra_inf}
This section provides comprehensive details on the specific training strategy and inference smoothing outlined in Section \textit{Training and Inference} in the main text. First, we elaborate on the training of the \textbf{lip attention} layers within our emotion-agnostic backbone. Second, we detail the \textbf{conflict-aware training strategy}, a critical mechanism that prevents emotional leakage from the reference image by utilizing mismatched reference-audio pairs, thereby forcing the model to rely solely on the audio signal for emotion modulation. Finally, we describe the \textbf{adaptive inter-frame smoothing strategy} employed during inference.

\noindent\textbf{Lip Attention for Emotion-agnostic Talking  Backbone.} To establish a solid foundation for speech-driven generation, we first train an emotion-agnostic backbone focused solely on precise lip synchronization. A key challenge in this stage is preventing the audio signal from influencing unrelated facial regions (e.g., blinking eyes, moving eyebrows) or the background. To address this, we introduce Lip Attention layers into the denoising U-Net. We use a binary spatial mask $M_\text{lip} \in \mathbb{R}^{h \times w}$ based on the facial landmarks of the source input image, where the mouth region is set to $1$ and other regions to $0$. This mask is downsampled to match the resolution of intermediate feature maps. Formally, let $Q$ be the visual queries from the noisy latent, and $K, V$ be the audio keys and values. The masked attention operation is defined as:
\begin{equation}
\text{Attention}(Q, K, V) = \text{Softmax}\left(\frac{Q K^\top}{\sqrt{d}} + \mathcal{M}\right) V,
\end{equation}
where $\mathcal{M}$ is the mask. If spatial position $(h, w)$ falls outside the mouth region (i.e., $M_{lip}^{(h,w)} = 0$), we set $\mathcal{M}^{(h,w)} = -\infty$, forcing the attention weights to zero. It is important to note that while Gaussian noise is added to the entire latent $z_t$ during training, the masking mechanism strictly blocks audio information from propagating to the upper face and background. Consequently, the denoising process for these regions relies exclusively on the spatial features provided by the reference net and the temporal context from self attention layers. This effectively "freezes" the non-mouth regions relative to the audio, ensuring that the backbone learns robust articulation.

\noindent\textbf{Conflict-aware Training Strategy.} A key challenge in emotional talking face generation is \emph{reference emotion leakage}: the generator may inadvertently copy the facial affect from the reference image $I_{\mathrm{ref}}$ instead of following the driving audio emotion, especially when strong identity guidance is provided by the reference network. To explicitly suppress this shortcut, we adopt an {conflict-aware training strategy} throughout the \textbf{GEM training stage}. Concretely, for each training sample, we construct a reference--audio pair $(\boldsymbol{I}_{\text{ref}}, \boldsymbol{A})$ such that the emotion label of the reference image/video clip $y_{\mathrm{ref}}$ and that of the driving audio $y_{a}$ are \emph{intentionally mismatched} with a controllable probability $p_{\mathrm{conf}}$. Specifically,
\[
\Pr\left(y_{\mathrm{ref}} \neq y_{a}\right)=p_{\mathrm{conf}}, \qquad 
\Pr\left(y_{\mathrm{ref}} = y_{a}\right)=1-p_{\mathrm{conf}}.
\]
In practice, we set $p_{\mathrm{conf}}=0.9$ unless otherwise specified. When sampling a mismatched pair, we keep the identity of $I_{\mathrm{ref}}$ unchanged and randomly draw an audio segment $A$ from a different emotion category than $y_{\mathrm{ref}}$ (while preserving the standard identity-disjoint split). This strategy forces the model to rely on the audio-derived implicit emotion representations $(f_l^{a}, f_e^{a})$ and the geometric priors $F_{\mathrm{geo}}$ for emotion modulation, rather than exploiting static affect cues from $I_{\mathrm{ref}}$. As a result, the generated expression becomes more faithful to the driving audio emotion and more robust under conflicting visual--audio affect conditions.

\noindent\textbf{Adaptive Inter-frame Smoothing Strategy.} Long-duration talking-face synthesis with a sliding-window pipeline may introduce temporal inconsistencies, typically manifested as background flickering and subtle facial jitter. A naive temporal average over all pixels often over-smooths critical motion cues (e.g., mouth articulation and eye dynamics), degrading lip-sync and expressiveness. To address this, we propose an \emph{adaptive}, \emph{region-aware} inter-frame smoothing strategy with a window size of $2$, which performs smoothing only when necessary and only on safe regions. Given two consecutive generated frames $I_{t-1}$ and $I_t$, we first compute region-wise appearance discrepancy to decide whether smoothing is required. Importantly, both the discrepancy computation and the subsequent fusion are restricted to the \emph{intersection} of the corresponding masks from the two frames, which avoids artifacts caused by mask boundary drift. The discrepancy thresholds used in our decisions are {not manually tuned}; instead, they are {estimated from statistics on real videos} by measuring the inter-frame appearance variation of background and facial regions under natural motion, and selecting thresholds according to the empirical distribution to reliably detect abnormal jitter. \textbf{Videos before and after smoothing are provided in the supplementary materials}.

\begin{itemize}[leftmargin=2em]
\item \emph{Background smoothing.} Let $M^{bg}_{t-1}$ and $M^{bg}_{t}$ denote the background masks for $I_{t-1}$ and $I_t$, respectively. We define the valid background region as the intersection $M^{bg}{\cap} = M^{bg}_{t-1} \cap M^{bg}_{t}$, and measure the background discrepancy only within $M^{bg}_{\cap}$. If the discrepancy exceeds the data-driven threshold, we replace pixels in $M^{bg}_{\cap}$ with the averaged result of the two frames, while leaving other regions unchanged, which selectively suppresses background flickering without affecting the foreground subject.

\item \emph{Face smoothing with organ protection.} Directly smoothing the entire face may blur the most important dynamic regions for perceptual realism and lip synchronization. Therefore, we construct a \emph{stable face} mask by removing the mouth and eye regions from the face area. Denote the resulting masks as $\tilde{M}^{face}_{t-1}$ and $\tilde{M}^{face}_{t}$, and define the valid region as $\tilde{M}^{face}_{\cap} = \tilde{M}^{face}_{t-1} \cap \tilde{M}^{face}_{t}$. We compute the discrepancy only on $\tilde{M}^{face}_{\cap}$. When it exceeds the data-driven threshold, we apply inter-frame averaging only within $\tilde{M}^{face}_{\cap}$, while preserving the mouth and eye regions to maintain accurate lip-sync and natural eye dynamics.

\item \emph{Optional intermediate-frame insertion.} In cases where the inter-frame discrepancy between $I_{t-1}$ and $I_t$ is exceptionally large (again determined by a real-data-based criterion), smoothing alone may still result in visible temporal jumps. We thus optionally insert an intermediate frame $I_{\text{mid}}$ between $I_{t-1}$ and $I_t$, computed by averaging the two frames restricted to the corresponding intersection masks. This insertion provides a gradual transition and further mitigates abrupt changes.

\item \emph{Stochastic action selection.} When smoothing is triggered, we randomly choose between (i) applying the region-restricted averaging on the current frame and (ii) inserting one intermediate frame between two adjacent frames, with equal probability ($50\%$-$50\%$). Qualitative demonstrations are provided in the supplementary videos, showing that this strategy improves temporal stability while avoiding over-smoothing of critical facial dynamics.
\end{itemize}

\section{Theoretical Analysis of Uncontrollable Magnitude and Geometric Awareness}
\label{app_sec_proof}
In this section, we provide a theoretical analysis to explain why implicit emotion representations suffer from \textbf{uncontrollable feature magnitude} and a \textbf{lack of geometric awareness}. We derive how these deficiencies negatively impact the attention mechanism in diffusion models and justify the necessity of our GEM module.

\noindent\textbf{Geometric Decomposition of Cross-Attention.} The core mechanism driving the facial generation is the Cross-Attention, where the noisy latent $z_t$ serves as the Query ($Q$) and the conditional emotion embedding serves as the Key ($K$). The attention weight $A_{i,j}$ for the $i$-th query token and $j$-th key token is computed as:
\begin{equation}
A_{i,j} \propto \exp\left( \frac{Q_i \cdot K_j^\top}{\sqrt{d}} \right) = \exp\left( \frac{\|Q_i\| \cdot \|K_j\| \cdot \cos\theta_{i,j}}{\sqrt{d}} \right),
\end{equation}
where $\theta_{i,j}$ is the angle between the feature vectors. This formula reveals that the influence of the condition is governed by two orthogonal components: (1) {direction ($\cos\theta$)}: Encodes the \textit{semantic category} (e.g., distinct muscle coordination patterns for "Happy" vs. "Sad"); (2) {magnitude ($\|K\|$)}: Encodes the \textit{signal strength} or activation intensity.

\noindent\textbf{Why Implicit Magnitude is Uncontrollable.} Implicit representations (extracted via V-AEP) inherently lack control over the magnitude $\|K\|$ due to the optimization objective of Contrastive Learning. The loss function used in V-AEP (and most implicit methods) is based on cosine similarity:
\begin{equation}
\begin{aligned}
\mathcal{L}_\text{con} &= -\log \frac{\exp(\text{sim}(z_i, z_p)/\tau)}{\sum \exp(\text{sim}(z_i, z_n)/\tau)}, \\
&\quad \text{where } \text{sim}(u, v) = \frac{u \cdot v}{\|u\| \|v\|}.
\end{aligned}
\end{equation}
The normalization term $\frac{1}{\|u\| \|v\|}$ projects all features onto a unit hypersphere. Consequently, the optimization gradient focuses purely on adjusting the angle $\theta$ to separate classes, while the magnitude $\|z\|$ becomes a free variable. This leads to radial invariance, where the learned magnitude $\|K\|$ is determined by random initialization or noise, rather than the physical intensity of the facial expression. The network has no incentive to encode "how wide the mouth opens" into the length of the vector.

\noindent\textbf{Lack of Geometric Awareness.} We define Geometric Awareness as the correlation between the feature norm and the physical deformation energy. 
\begin{itemize}[leftmargin=2em]
    \item \textbf{Explicit Geometric Priors (Blendshapes):} Let $g \in \mathbb{R}^{45}$ be the blendshape coefficients. The norm $\|g\|_2$ is physically grounded: a larger $\|g\|_2$ directly corresponds to larger muscle displacements (e.g., a wider jaw opening).
    \item \textbf{Implicit Representations:} Let $f_{imp}$ be the implicit feature. Due to the radial invariance described above, $\|f_\text{imp}\|$ is uncorrelated with physical deformation. A "screaming" face and a "smiling" face might have similar implicit feature norms, or worse, a neutral face might have a larger norm due to noise. This disconnect means implicit features lack the geometric awareness required to tell the generator how much to deform the mesh.
\end{itemize}

\noindent\textbf{Consequence: Dispersed Attention and Loss of Control.} The fundamental issue arises from the dense interaction between the noisy latent pixels ($Q \in \mathbb{R}^{B\times N\times 4096 \times d}$) and the conditional audio tokens ($K \in \mathbb{R}^{B\times N \times32 \times d}$). Ideally, the attention mechanism should assign high scores precisely to the pixels corresponding to active muscles. However, in the implicit baseline, the ungrounded feature magnitude acts as a chaotic scaling factor $\eta$ that corrupts the semantic correlation. Mathematically, we can model the attention logit $Z_{i,j}$ (before Softmax) as the product of a semantic component and a noise component:
\begin{equation}
Z_{i,j} = \underbrace{\eta_j}_{\text{Chaotic Magnitude}} \cdot \underbrace{\mathcal{C}_{i,j}}_{\text{Semantic Direction}},
\end{equation}
where $\mathcal{C}_{i,j}$ represents the correct geometric alignment (e.g., audio token $j$ aligning with lip pixels $i$), but $\eta_j$ is a random variable with high variance. This multiplicative noise disrupts the generation in two ways: first, a small $\eta_j$ prevents the target muscles from being strongly activated ("weak drawing force"); second, a randomly large $\eta_j$ allows background noise to dominate, causing the attention to drift to irrelevant regions.

This chaotic interaction inevitably leads to a \textbf{dispersed attention map}, where the probability mass is smeared across the face rather than focused on the target muscles. Consequently, the most direct and critical failure of this structural blindness is the \textcolor{red}{\textbf{inability to continuously control emotional intensity}}. Because the "force" of the condition is randomized, the model cannot distinguish between varying degrees of emotion (e.g., a subtle smirk vs. a broad grin) and defaults to a generic "Average Face", failing to produce the smooth, continuous transitions required for realistic, high-fidelity expression.

\section{Efficiency Analysis}
\begin{table}[t]
\centering
\caption{Efficiency analysis of different settings of our proposed method. It shows, from top to bottom, the inference time and inference GPU memory as we progressively add the components we propose.}
\label{tab:efficiency}
\resizebox{0.98\columnwidth}{!}{
\begin{tabular}{l|cc}
\toprule
\textbf{Method} & {GPU memory (GB)} & {Time (sec)} \\ 
\midrule
Inference: Backbone 
& 9.53 & 12.36 \\

Inference: + V-AEP 
& 9.57 & 13.91 \\

Inference: + D-GPG 
& 10.31 & 14.67 \\

\midrule
Inference: + Smoothing 
& 10.31 & 14.67 \\ 
\bottomrule

\end{tabular}
}
\end{table}
Table \ref{tab:efficiency} presents the efficiency analysis of our method across various stages. The inference time of our emotionless talking backbone is 12.36 seconds, with a GPU memory usage of 9.53GB. In comparison, incrementally incorporating our V-AEP and D-GPG modules to achieve emotion-controllable talking results in only a 0.78GB increase in GPU memory and a 2.31-second increase in inference time. Furthermore, our Adaptive Inter-frame Smoothing Strategy does not increase GPU memory or inference time, as the smoothing process involves no model computations and can be executed alternately with inference.

 \section{Limitations}
 Our method integrates explicit and implicit representations to achieve continuous emotion-controllable talking faces, but limitations highlight future research directions: (1) \textbf{Multi-face perspective}. This method currently does not support multi-angle face talking generation, as the dataset we selected lacks multi-angle samples. The primary focus of our study was to validate the effectiveness of combining implicit and explicit representations. Incorporating multi-angle capabilities in future work will significantly enhance the practical applicability of this technology. (2) \textbf{Full-Body Human Talking Generation}. Current research remains primarily focused on full-body human talking generation. Integrating fine-grained emotional features—such as expressive gestures, dynamic body postures, and nuanced movement patterns—into the entire human model would significantly enhance the realism and expressiveness of digital humans, making them more lifelike and contextually engaging. (3) \textbf{Computational Efficiency}. Optimizing the computational efficiency of diffusion models is critical for practical deployment. Investigating lightweight architectures, quantization techniques, and model distillation methods holds promise for enabling real-time applications while minimizing computational resource consumption and maintaining performance integrity. 
 
 Our supplementary material includes experimental video results demonstrating the effectiveness of the proposed approach. Figure \ref{app_fig:dataset_result} presents the performance of our method across multiple datasets, including out-of-domain data, to showcase its generalization capability.

\begin{figure*}[h]
  \centering
  \includegraphics[width=0.70\textwidth, trim=0 0 0 0, clip]{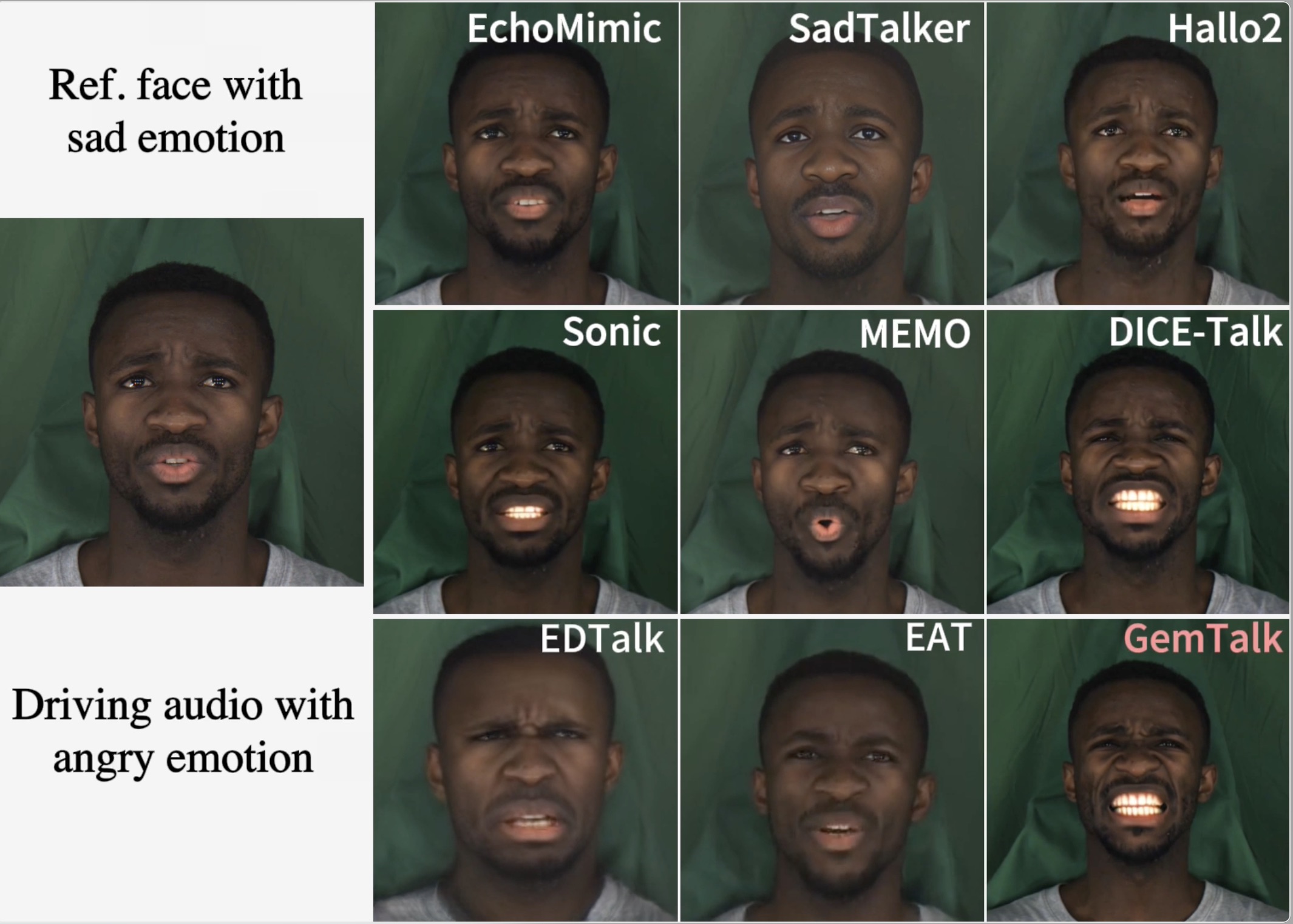}
  \caption{Qualitative comparison with SOTA methods on MEDA with Angry expression.}
  \label{app_fig:compare}
\end{figure*}

\begin{figure*}[h]
  \centering
  \includegraphics[width=0.70 \textwidth, trim=0 0 0 0, clip]{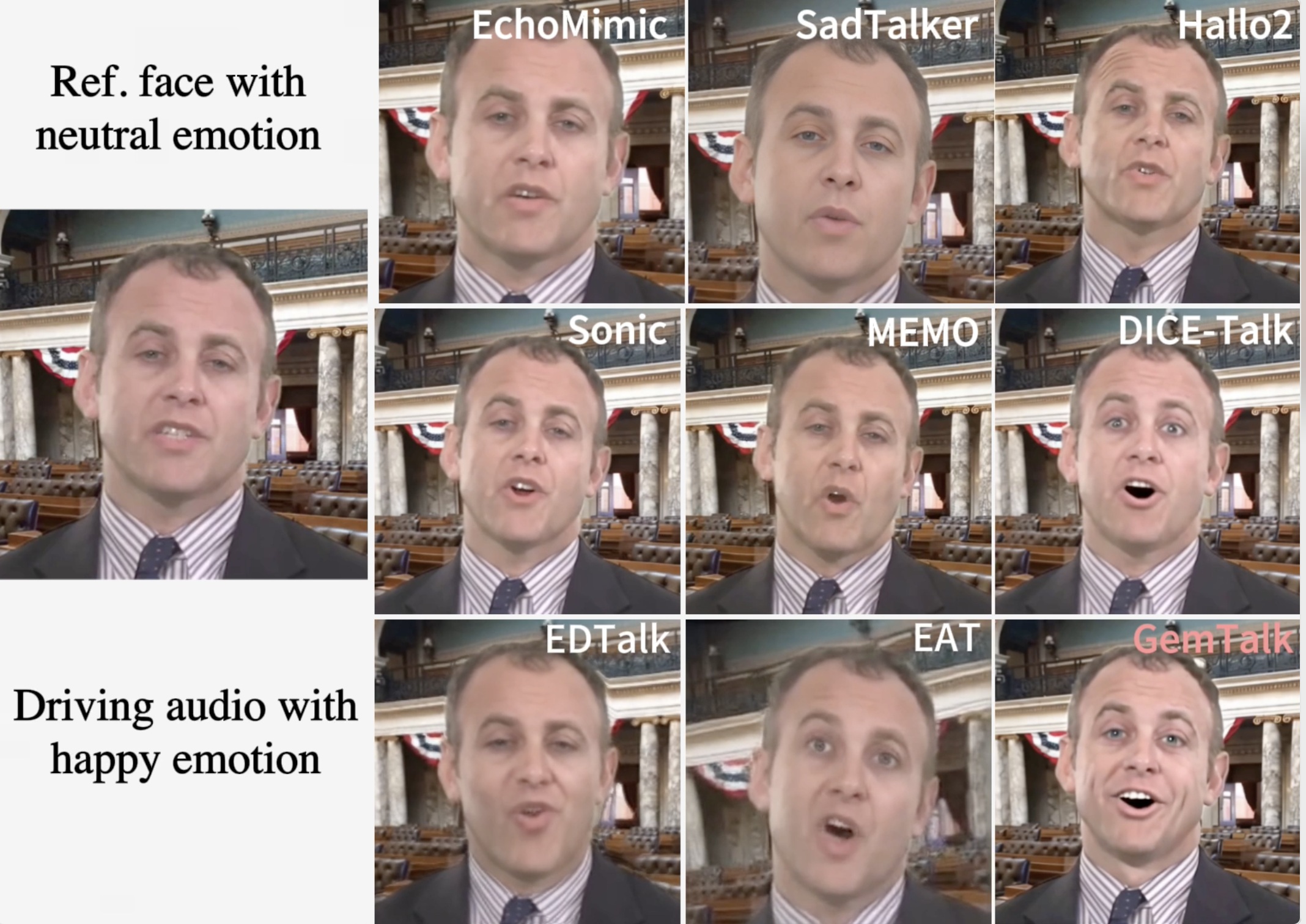}
  \caption{Qualitative comparison with SOTA methods on HDTF with Happy expression.}
  \label{app_fig:dataset_result}
\end{figure*}

\begin{figure*}[h]
  \centering
  \includegraphics[width=0.95 \textwidth, trim=0 0 0 0, clip]{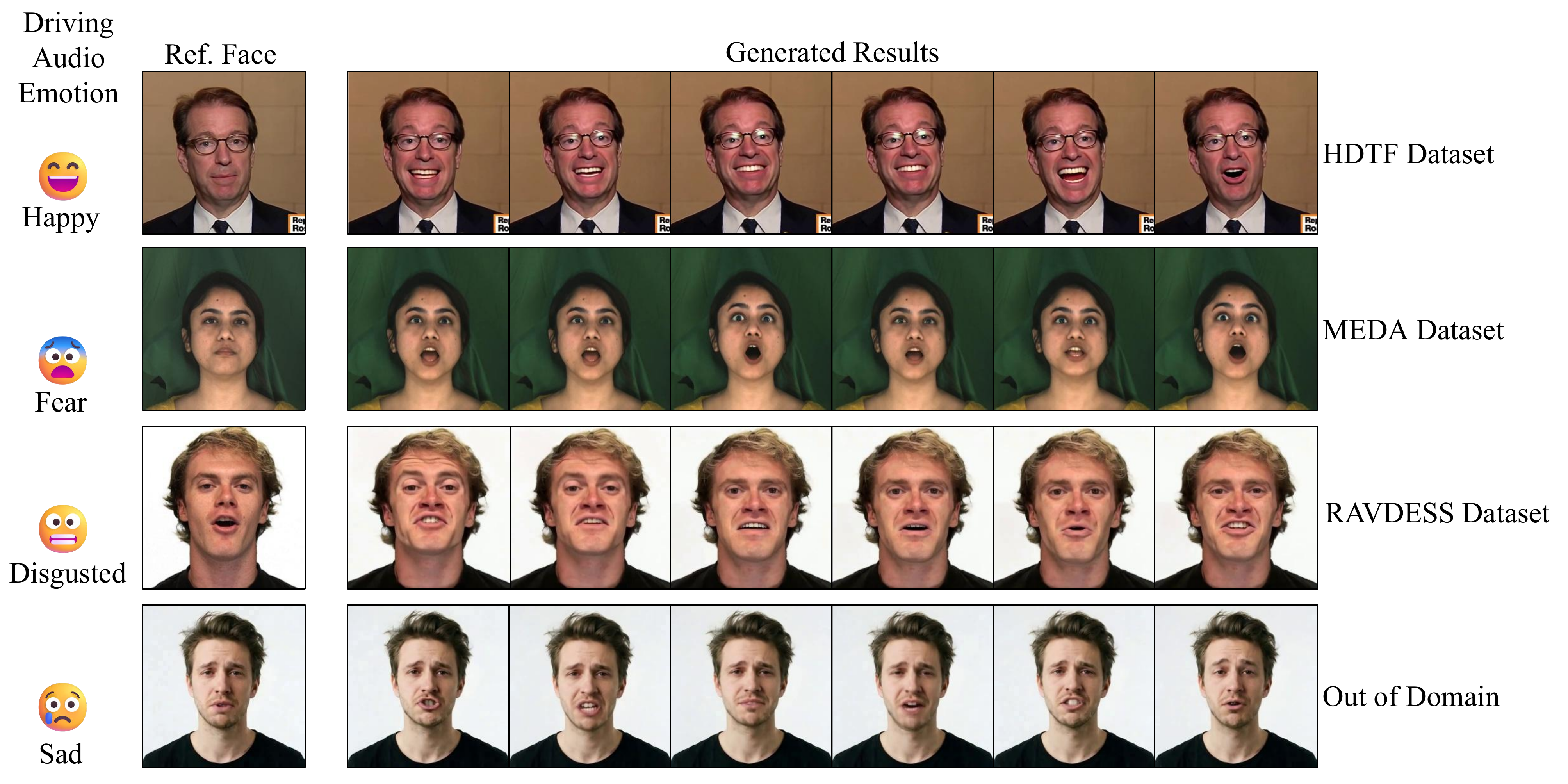}
  \caption{Our GemTalk facial emotion generation performance on different datasets.}
  \label{app_fig:dataset_result}
\end{figure*}


\end{document}